\documentclass{article} 

\usepackage{amsmath, amssymb, amsthm}
\usepackage{mathtools}
\usepackage{bbm}
\usepackage{dsfont}

\usepackage[usenames,dvipsnames]{xcolor}

\usepackage{graphicx}
\usepackage[labelfont=bf]{caption}
\usepackage[format=hang]{subcaption}

\usepackage{booktabs, array}
\usepackage{multirow}

\usepackage[algoruled]{algorithm2e}
\usepackage{listings}
\usepackage{fancyvrb}
\fvset{fontsize=\small}

\usepackage{natbib}
\usepackage[colorlinks,linktoc=all,pagebackref=true]{hyperref}
\usepackage[hypcap=true]{caption}
\hypersetup{citecolor=NavyBlue}
\hypersetup{linkcolor=NavyBlue}
\hypersetup{urlcolor=NavyBlue}
\usepackage[nameinlink,capitalise]{cleveref}
\creflabelformat{equation}{#2\textup{#1}#3}  

\lstdefinestyle{mystyle}{
    commentstyle=\color{OliveGreen},
    numberstyle=\tiny\color{black!60},
    stringstyle=\color{BrickRed},
    basicstyle=\ttfamily\scriptsize,
    breakatwhitespace=false,
    breaklines=true,
    captionpos=b,
    keepspaces=true,
    numbers=none,
    numbersep=5pt,
    showspaces=false,
    showstringspaces=false,
    showtabs=false,
    tabsize=2
}
\newsavebox\CBox 
\newcommand{\spm}[1]{\scriptstyle{\pm#1}}
\def\textBF#1{\sbox\CBox{#1}\resizebox{\wd\CBox}{\ht\CBox}{\textbf{#1}}}

\usepackage[acronym,nowarn,section,nogroupskip,nonumberlist]{glossaries}
\glsdisablehyper{}

\newacronym[\glslongpluralkey={Gaussian Processes}]{gp}{\textsc{gp}}{Gaussian Process}
\newacronym[\glslongpluralkey={Conditional Neural Processes}]{cnp}{\textsc{cnp}}{Conditional Neural Process}
\newacronym[\glslongpluralkey={Neural Processes}]{np}{\textsc{np}}{Neural Process}
\newacronym[\glslongpluralkey={Neural Process Families}]{npf}{\textsc{npf}}{Neural Process Family}
\newacronym[\glslongpluralkey={Attentive Neural Processes}]{anp}{\textsc{anp}}{Attentive Neural Process}
\newacronym[\glslongpluralkey={Conditional Attentive Neural Processes}]{canp}{\textsc{canp}}{Conditional Attentive Neural Process}
\newacronym[\glslongpluralkey={Convolutional Conditional Neural Processes}]{convcnp}{\textsc{c}onv\textsc{cnp}}{Convolutional Conditional Neural Processes}
\newacronym[\glslongpluralkey={Convolutional Neural Processes}]{convnp}{\textsc{c}onv\textsc{np}}{Convolutional Neural Processes}
\newacronym[\glslongpluralkey={Bootstrapping Neural Processes}]{bnp}{\textsc{bnp}}{Bootstrapping Neural Process}
\newacronym[\glslongpluralkey={Neural Bootstrapping Neural Processes}]{neubnp}{\textsc{n}eu\textsc{bnp}}{Neural Bootstrapping Neural Process}
\newacronym[\glslongpluralkey={Martingale Posterior Neural Processes}]{mpnp}{\textsc{mpnp}}{Martingale Posterior Neural Process}
\newacronym[\glslongpluralkey={Bootstrapping Attentive Neural Processes}]{banp}{\textsc{banp}}{Bootstrapping Attentive Neural Process}
\newacronym[\glslongpluralkey={Neural Bootstrapping Attentive Neural Processes}]{neubanp}{\textsc{n}eu\textsc{banp}}{Neural Bootstrapping Attentive Neural Process}
\newacronym[\glslongpluralkey={Martingale Posterior Attentive Neural Processes}]{mpanp}{\textsc{mpanp}}{Martingale Posterior Attentive Neural Process}
\newacronym[\glslongpluralkey={Multi-Layer Perceptrons}]{mlp}{\textsc{mlp}}{Multi-Layer Perceptron}
\newacronym{elbo}{\textsc{elbo}}{Evidence Lower BOund}
\newacronym{cid}{c.i.d.}{conditionally identically distributed}
\newacronym{mab}{\textsc{mab}}{Multihead Attention Block}
\newacronym{isab}{\textsc{isab}}{Induced Self-Attention Block}

\newcommand{\enc}{\text{enc}}

\newcommand{\gen}{\text{gen}}

\newcommand{\In}{\text{in}}
\newcommand{\hid}{\text{hid}}
\newcommand{\head}{\text{head}}
\newcommand{\out}{\text{out}}
\newcommand{\MLP}{\text{MLP}}
\newcommand{\Lin}{\text{Lin}}
\newcommand{\LN}{\text{LN}}
\newcommand{\concat}{\text{concat}}
\newcommand{\softmax}{\text{softmax}}
\newcommand{\ISAB}{\text{ISAB}}
\newcommand{\MAB}{\text{MAB}}
\newcommand{\MHA}{\text{MHA}}
\newcommand{\ReLU}{\text{ReLU}}
\newcommand{\spl}{\text{split}}
\newcommand{\SA}{\text{SA}}

\newcommand{\bof}{\mathbf{f}}

\newcommand{\calE}{{\mathcal{E}}}

\newcommand{\calL}{{\mathcal{L}}}

\newcommand{\calN}{{\mathcal{N}}}

\newcommand{\calT}{{\mathcal{T}}}

\newcommand{\calX}{{\mathcal{X}}}
\newcommand{\calY}{{\mathcal{Y}}}
\newcommand{\calZ}{{\mathcal{Z}}}

\newcommand{\bbE}{\mathbb{E}}

\newcommand{\bbR}{\mathbb{R}}

\theoremstyle{plain}

\theoremstyle{definition}

\theoremstyle{remark}

\newcommand{\dee}{\mathrm{d}}

\DeclareMathOperator*{\argmin}{arg\,min}

\newcommand{\iidsim}{\overset{\mathrm{i.i.d.}}{\sim}}

\newcommand{\KL}{D_{\mathrm{KL}}}

\newcommand{\din}{{d_{\text{in}}}}
\newcommand{\dout}{{d_{\text{out}}}}

\def\[#1\]{\begin{align}#1\end{align}}

\usepackage{iclr2023_conference,times}

\title{Martingale Posterior Neural Processes}

\author{Hyungi~Lee$^{1}$, Eunggu~Yun$^{1}$, Giung~Nam$^{1}$, Edwin~Fong$^{2}$, Juho~Lee$^{1,3}$ \\ 
\vspace{-0.1in}
\\
$^1$KAIST, $^2$Novo Nordisk, $^3$AITRICS \\
\footnotesize
\texttt{$^1$\{lhk2708,\,eunggu.yun,\,giung,\,juholee\}@kaist.ac.kr, $^2$chef@novonordisk.com}
}

\iclrfinalcopy 

\begin{document}

\maketitle

\begin{abstract}
A \gls{np} estimates a stochastic process implicitly defined with neural networks given a stream of data, rather than pre-specifying priors already known, such as Gaussian processes. An ideal \gls{np} would learn everything from data without any inductive biases, but in practice, we often restrict the class of stochastic processes for the ease of estimation. One such restriction is the use of a finite-dimensional latent variable accounting for the uncertainty in the functions drawn from \glspl{np}. Some recent works show that this can be improved with more ``data-driven’’ source of uncertainty such as bootstrapping. In this work, we take a different approach based on the martingale posterior, a recently developed alternative to Bayesian inference. For the martingale posterior, instead of specifying prior-likelihood pairs, a predictive distribution for future data is specified. Under specific conditions on the predictive distribution, it can be shown that the uncertainty in the generated future data actually corresponds to the uncertainty of the implicitly defined Bayesian posteriors. Based on this result, instead of assuming any form of the latent variables, we equip a \gls{np} with a predictive distribution implicitly defined with neural networks and use the corresponding martingale posteriors as the source of uncertainty. The resulting model, which we name as \gls{mpnp}, is demonstrated to outperform baselines on various tasks.

\end{abstract}

\section{Introduction}
\label{main:sec:introduction}

\glsresetall

A \gls{np}~\citep{garnelo2018conditional,garnelo2018neural} meta-learns a stochastic process describing the relationship between inputs and outputs in a given data stream, where each task in the data stream consists of a meta-training set of input-output pairs and also a meta-validation set. The \gls{np} then defines an implicit stochastic process whose functional form is determined by a neural network taking the meta-training set as an input, and the parameters of the neural network are optimized to maximize the predictive likelihood for the meta-validation set. This approach is philosophically different from the traditional learning pipeline where one would first elicit a stochastic process from the known class of models (e.g., \glspl{gp}) and hope that it describes the data well. An ideal \gls{np} would assume minimal inductive biases and learn as much as possible from the data. In this regard, \glspl{np} can be framed as a ``data-driven'' way of choosing proper stochastic processes.

 An important design choice for a \gls{np} model is how to capture the uncertainty in the random functions drawn from stochastic processes. When mapping the meta-training set into a function, one might employ a deterministic mapping as in \citet{garnelo2018conditional}. However, it is more natural to assume that there may be multiple plausible functions that might have generated the given data, and thus encode the functional (epistemic) uncertainty as a part of the \gls{np} model. \citet{garnelo2018neural} later proposed to map the meta-training set into a fixed dimensional \emph{global latent variable} with a Gaussian posterior approximation. While this improves upon the vanilla model without such a latent variable~\citep{le2018empirical}, expressing the functional uncertainty only through the Gaussian approximated latent variable has been reported to be a bottleneck~\citep{louizos2019functional}. To this end, \citet{lee2020bootstrapping} and \citet{lee2022neural} propose to apply bootstrap to the meta-training set to use the uncertainty arising from the population distribution as a source for the functional uncertainty.

In this paper, we take a rather different approach to define the functional uncertainty for \glspl{np}. Specifically, we utilize the martingale posterior distribution~\citep{fong2021martingale}, a recently developed alternative to conventional Bayesian inference. In the martingale posterior, instead of eliciting a likelihood-prior pair and inferring the Bayesian posterior, we elicit a joint predictive distribution on future data given observed data. Under suitable conditions on such a predictive distribution, it can be shown that the uncertainty due to the generated future data indeed corresponds to the uncertainty of the Bayesian posterior. Following this, we endow a \gls{np} with a joint predictive distribution defined through neural networks and derive the functional uncertainty as the uncertainty arising when mapping the randomly generated future data to the functions. Compared to the previous approaches of either explicitly positing a finite-dimensional variable encoding the functional uncertainty or deriving it from a population distribution, our method makes minimal assumptions about the predictive distribution and gives more freedom to the model to choose the proper form of uncertainty solely from the data. Due to the theory of martingale posteriors, our model guarantees the existence of the martingale posterior corresponding to the valid Bayesian posterior of an implicitly defined parameter. 
Furthermore, working in the space of future observations allows us to incorporate the latent functional uncertainty path with deterministic path in a more natural manner.

We name our extension of \glspl{np} with the joint predictive generative models as the \gls{mpnp}. Throughout the paper, we propose an efficient neural network architecture for the generative model that is easy to implement, flexible, and yet guarantees the existence of the martingale posterior. We also propose a training scheme to stably learn the parameters of \glspl{mpnp}. Using various synthetic and real-world regression tasks, we demonstrate that \gls{mpnp} significantly outperforms the previous \gls{np} variants in terms of predictive performance.

\section{Background}
\label{main:sec:background}
\subsection{Settings and notations}

Let $\calX = \bbR^{\din}$ be an input space and $\calY = \bbR^\dout$ be an output space. 
We are given a set of \emph{tasks} drawn from an (unknown) task distribution, $\tau_1, \tau_2, \dots \iidsim p_\text{task}(\tau)$. 
A task $\tau$ consists of a dataset $Z$ and an index set $c$, where $Z = \{z_i\}_{i=1}^n$ with each $z_i = (x_i, y_i) \in \calX \times \calY$ is a pair of an input and an output. We assume $Z$ are i.i.d. conditioned on some function $f$. The index set $c \subsetneq [n]$ where $[n] := \{1,\dots, n\}$ defines the \emph{context set} $Z_c = \{z_i\}_{i\in c}$. The \emph{target set} $Z_t$ is defined similarly with the index $t := [n]\setminus c$.


\subsection{Neural process families}

Our goal is to train a class of random functions $f: \calX \to \calY$ that can effectively describe the relationship between inputs and outputs included in a set of tasks. Viewing this as a meta-learning problem, for each task $\tau$, we can treat the context $Z_c$ as a meta-train set and target $Z_t$ as a meta-validation set. We wish to meta-learn a mapping from the context $Z_c$ to a random function $f$ that recovers the given context $Z_c$ (minimizing meta-training error) and predicts $Z_t$ well (minimizing meta-validation error). Instead of directly estimating the infinite-dimensional $f$, we learn a mapping from $Z_c$ to a predictive distribution for finite-dimensional observations,
\[
p(Y | X, Z_c) = \int \bigg[\prod_{i\in c} p(y_i | f, x_i) \prod_{i\in t} p(y_i | f, x_i)\bigg] p(f|Z_c) \dee f,
\]
where we are assuming the outputs $Y$ are independent given $f$ and $X$. We further restrict ourselves to simple heteroscedastic Gaussian measurement noises,
\[
p(y|f, x) = \calN(y | \mu_\theta(x), \sigma^2_\theta(x)I_{\dout}),
\]
where $\mu_\theta: \calX \to \calY$ and $\sigma_\theta^2: \calX \to \bbR_+$ map an input to a mean function value and corresponding variance, respectively. $\theta \in \bbR^{h}$ is a parameter indexing the function $f$, and thus the above predictive distribution can be written as
\[
p(Y | X, Z_c) = \int 
\bigg[\prod_{i\in [n]} \calN(y_i|\mu_\theta(x_i), \sigma_\theta^2(x_i) I_\dout)\bigg] p(\theta|Z_c) \dee \theta.
\]
A \gls{np} is a parametric model which constructs a mapping from $Z_c$ to $\theta$ as a neural network. The simplest version, \gls{cnp}~\citep{garnelo2018conditional}, assumes a deterministic mapping from $Z_c$ to $\theta$ as
\[
p(\theta|Z_c) = \delta_{r_c}(\theta), \quad r_c = f_\text{enc}(Z_c ; \phi_\text{enc}),
\]
where $\delta_{a}(x)$ is the Dirac delta function (which gives zero if $x\neq a$ and $\int\delta_a(x) \dee x=1$) and $f_\text{enc}$ is a \emph{permutation-invariant} neural network taking sets as inputs~\citep{zaheer2017deep}, parameterized by $\phi_\text{enc}$. Given a summary $\theta= r_c$ of a context $Z_c$, the \gls{cnp} models the mean and variance functions $(\mu, \sigma^2)$ as
\[
(\mu_\theta(x), \log \sigma_\theta(x)) = f_\text{dec}(x, r_c ; \phi_\text{dec}),
\]
where $f_\text{dec}$ is a feed-forward neural network parameterized by $\phi_\text{dec}$. Here the parameters $(\phi_\text{enc}, \phi_\text{dec})$ are optimized to maximize the expected predictive likelihood over tasks, $\bbE_{\tau}[\log p(Y|X, Z_c)]$.

Note that in the \gls{cnp}, the mapping from $Z_c$ to $\theta$ is deterministic, so it does not consider \emph{functional uncertainty} or epistemic (model) uncertainty. To resolve this, \citet{garnelo2018neural} proposed \gls{np} which learns a mapping from an arbitrary subset $Z' \subseteq Z$ to a variational posterior $q(\theta|Z')$ approximating $p(\theta|Z')$ under an implicitly defined prior $p(\theta)$: 
\[
(m_{Z'}, \log s_{Z'}) = f_\text{enc}(Z'; \phi_\text{enc}), \quad p(\theta|Z') \approx q(\theta|Z') := \calN(\theta | m_{Z'}, s^2_{Z'}I_h).
\]
With $f_\text{enc}$, the \gls{elbo} for the predictive likelihood is written as
\[
\log p(Y|X,Z_c) &\geq \sum_{i\in[n]} \bbE_{q(\theta|Z)}[\log \calN(y_i|\mu_\theta(x_i), \sigma_\theta^2(x_i)I_\dout)] - \KL[q(\theta|Z)\Vert p(\theta|Z_c)] \nonumber\\
&\approx
\sum_{i\in[n]} \bbE_{q(\theta|Z)}[\log \calN(y_i|\mu_\theta(x_i), \sigma_\theta^2(x_i)I_\dout)] - \KL[q(\theta|Z)\Vert q(\theta|Z_c)].
\]
An apparent limitation of the \gls{np} is that it assumes a uni-modal Gaussian distribution as an approximate posterior for $q(\theta|Z_c)$. Aside from the limited flexibility, it does not fit the motivation of \glspl{np} trying to learn as much as possible in a data-driven manner, as pre-specified parametric families are used.  

There have been several improvements over the vanilla \glspl{cnp} and \glspl{np}, either by introducing attention mechanism~\citep{vaswani2017attention} for $f_\text{enc}$ and $f_\text{dec}$~\citep{kim2018attentive}, or using advanced functional uncertainty modeling~(\citealp{lee2020bootstrapping}; \citealp{lee2022neural}). We provide a detailed review of the architectures for such variants in \cref{app:sec:architectures}. Throughout the paper, we will refer to this class of models as \gls{npf}.

\subsection{Martingale Posterior Distributions}

The martingale posterior distribution \citep{fong2021martingale} is a recent generalization of Bayesian inference which reframes posterior uncertainty on parameters as {predictive} uncertainty on the unseen population conditional on the observed data. Given observed samples $Z = \{z_i\}_{i=1}^n$  i.i.d. from the sampling density $p_0$, one can define the parameter of interest as a functional of $p_0$, that is
$$
\theta_0 = \theta(p_0) =  \argmin_\theta\int \ell(z,\theta)\, p_0(dz), 
$$
where $\ell$ is a loss function. For example, $\ell(z,\theta) =  (z-\theta)^2$ would return $\theta_0$ as the mean, and $\ell(z,\theta)= - \log p(z \mid \theta)$ would return the KL minimizing parameter between $p(\cdot \mid \theta)$ and $p_0$.

The next step of the martingale posterior is to construct a \emph{joint} predictive density on $Z' = \{z_i\}_{i=n+1}^N$ for some large $N$, which we write as $p(Z' \mid Z)$. In a similar fashion to a bootstrap, one can imagine drawing $Z' \sim p(Z' \mid Z)$, then computing $\theta(g_N)$ where $g_N(z) = \frac{1}{N} \sum_{i=1}^N \delta_{z_i} (z)$. 
The predictive uncertainty in $Z'$ induces uncertainty in $\theta(g_N)$ conditional on $Z$. The key connection is that  if $p(Z' \mid Z)$ is the Bayesian joint  posterior predictive density, and $\ell = - \log p(z \mid \theta)$, then $\theta(g_N)$ is distributed according to the Bayesian posterior $\pi(\theta \mid Z)$ as $N \to \infty$, under weak conditions. In other words, posterior uncertainty in $\theta$ is equivalent to predictive uncertainty in $\{z_i\}_{i=n+1}^\infty$. 

\cite{fong2021martingale} specify more general  $p(Z' \mid Z)$ directly beyond the Bayesian posterior predictive, and define the (finite) martingale posterior 
as 
$\pi_N(\theta \in A \mid Z) = \int \mathbbm{1}(\theta(g_N) \in A) \, p(dZ' \mid Z)$. In particular, the joint predictive density can be factorized into a sequence of 1-step-ahead predictives, 
$
p(Z' \mid Z) = \prod_{i=n+1}^N p(z_i \mid z_{1:i-1}),
$
and the sequence $\{p(z_i \mid z_{1:i-1})\}_{n+1}^N$ is elicited directly, removing the need for the likelihood and prior. Hyperparameters for the sequence of predictive distributions can be fitted in a data-driven way by maximizing 
$$\log p(Z) = \sum_{i=1}^n \log p(z_i \mid z_{1:i-1}),$$ 
which is analogous to the log marginal likelihood.  \cite{fong2021martingale} requires the sequence of predictives to be \gls{cid}, which is a martingale condition on the sequence of predictives that ensures $g_N$ exists almost surely. The Bayesian posterior predictive density is a special case, as exchangeability of $p(Z' \mid Z)$ implies the sequence of predictives is \gls{cid} In fact, De Finetti's theorem \citep{de1937prevision} guarantees that any exchangeable joint density implies an underlying likelihood-prior form, but specifying the predictive density directly can be advantageous. It allows for  easier computation, as we no longer require posterior approximations, and it also widens the class of available nonparametric predictives which we will see shortly.  

\subsection{Exchangeable Generative Models}\label{main:subsec:exchangeable}

To construct a martingale posterior, we can either specify a sequence of one-step predictive distributions or the joint predictive density distribution directly, as long as the \gls{cid} condition is satisfied.  Here, we opt to specify an exchangeable $p(Z' \mid Z)$ directly, which then implies the required \gls{cid} predictives. We now briefly review exchangeable generative models which can be used to specify the exchangeable joint predictive.  For a set of random variables $Z = \{z_i\}_{i=1}^n$ with each $z_i\in \calZ = \bbR^{d}$, we say the joint distribution $p(Z)$ is \emph{exchangeable} if it is invariant to the arbitrary permutation of the indices, that is, $p(Z) = p(\pi\cdot Z)$ for any permutation $\pi$ of $[n]$. A simple way to construct such exchangeable random variables is to use a \emph{permutation-equivariant mapping}. A mapping $\bof: \calZ^n\to\calZ^n$ is permutation equivariant if $\bof(\pi\cdot Z) = \pi\cdot\bof(Z)$ for any $\pi$. Given $\bof$, we can first generate i.i.d. random variables and apply $\bof$ to construct a potentially correlated but exchangeable set of random variables $Z$ as follows:
\[
\calE := \{\varepsilon_i\}_{i=1}^n \iidsim p_0, \quad Z = \bof(\calE).
\]
For $\bof$, we employ the modules introduced in \citet{lee2019set}. Specifically, we use a permutation equivariant module called  \gls{isab}. An \gls{isab} mixes input sets through a learnable set of parameters called \emph{inducing points} via \glspl{mab}~\citep{vaswani2017attention,lee2019set}. 
\[
\textsc{isab}(\calE) = \textsc{mab}(\calE, H) \in \bbR^{n\times d}\text{ where } H = \textsc{mab}(I, \calE)\in \bbR^{m\times d}.
\]
Here, $I \in \bbR^{m\times d}$ is a set of $m$ inducing points and $\textsc{mab}(\cdot,\cdot)$ computes attention between two sets.
The time-complexity of an \gls{isab} is $O(nm)$, scales linear with input set sizes.

\section{Methods}
\label{main:sec:methods}
In this section, we present a novel extension of \gls{npf} called \glspl{mpnp}. The main idea is to elicit joint predictive distributions that are constructed with equivariant neural networks instead of assuming priors for $\theta$, and let the corresponding martingale posterior describe the functional uncertainty in the \glspl{np}. We describe how we construct a \gls{mpnp} in \cref{main:subsec:mpnp} and train it in \cref{main:subsec:training}.


\subsection{Martingale Posterior Neural Processes}\label{main:subsec:mpnp}

Recall that the functional uncertainty in a \gls{np} is encoded in a parameter $\theta$. Rather than learning an approximate posterior $q(\theta|Z_c)$, we introduce a joint predictive $p(Z'|Z_c ; \phi_\text{pred})$ generating a \emph{pseudo context set} $Z' = \{z'_i\}_{i=1}^{N-|c|}$ of size $(N-|c|) \geq 1$. 
Having generated a pseudo context, we combine with the existing context $Z_c$, and construct the empirical density as
\[
g_N(z) = \frac{1}{N}\bigg(\sum_{i\in c}\delta_{z_i}(z) + \sum_{i=1}^{N-|c|} \delta_{z'_i}(z)\bigg).
\]
Given $g_N$, the estimate of the function parameter $\theta$ is then recovered as 
\[
\label{eq:recovering_theta}
\theta(g_N) := \argmin_\theta \int \ell(z, \theta) g_N(dz),
\]
where in our case we simply choose $\ell(z,\theta) := -\log \calN(y|\mu_\theta(x), \sigma^2_\theta(x)I_\dout)$. The uncertainty in $\theta(g_N)$ is thus induced by the uncertainty in the generated pseudo context $Z'$.

\paragraph{Amortization}
The procedure of recovering $\theta$ via \cref{eq:recovering_theta} would originally require iterative optimization process except for simple cases. Fortunately, in our case, we can \emph{amortize} this procedure, thanks to the mechanism of \glspl{cnp} amortizing the inference procedure of estimating $\theta$ from the context. Given $Z_c$, a \gls{cnp} learns an encoder producing $\theta$ that is trained to maximize the expected likelihood. That is,
\[
 \tilde\theta(Z_c) = f_\text{enc}(Z_c;\phi_\text{enc}), \quad \tilde\theta(Z_c) \approx \argmin_{\theta} \int \ell(z,\theta) g_c(dz),
\]
where $g_c$ is the empirical density of $Z_c$. Hence, given $Z'$ and $Z_c$, we can just input $Z'\cup Z_c$ into $f_\text{enc}$ and use the output $\tilde\theta(Z'\cup Z_c)$ as a proxy for $\theta(g_N)$. Compared to exactly computing $\theta(g_N)$, obtaining $\tilde\theta(Z'\cup Z_c)$ requires a single forward pass through $f_\text{enc}$, which scales much better with $N$. Moreover, computation for multiple $Z'$ required for bagging can easily be parallelized. 

\paragraph{Specifying the joint predictives}
We construct the joint predictives $p(Z'|Z_c;\phi_\text{pred})$ with a neural network. Other than the requirement of $Z'$ being exchangeable (and thus \gls{cid}), we give no inductive bias to $p(Z'|Z_c;\phi_\text{pred})$ and let the model learn $\phi_\text{pred}$ from the data. We thus use the exchangeable generative model described in \cref{main:subsec:exchangeable}. Specifically, to generate $Z'$, we first generate $\calE = \{\varepsilon_i\}_{i=1}^n$ from some distribution (usually chosen to be a unit Gaussian $\calN(0, I_d)$), and pass them through an equivariant \gls{isab} block to form $Z'$. To model the conditioning on $Z_c$, we set the inducing point in the \gls{isab} as a transform of $Z_c$. That is, with an arbitrary feed-forward neural network $h$,
\[
\textsc{isab}(\calE) = \textsc{mab}(\calE, H), \quad H = \textsc{mab}( h(Z_c), \calE ),
\]
where $h(Z_c) = \{h(z_i)\}_{i\in c}$. 
The resulting model is an implicit generative model~\citep{mohamed2016learning} in a sense that we can draw samples from it but cannot evaluate likelihoods.

\paragraph{Generating Representations} 
\begin{figure}[t]
    \centering
    \includegraphics[width=\linewidth]{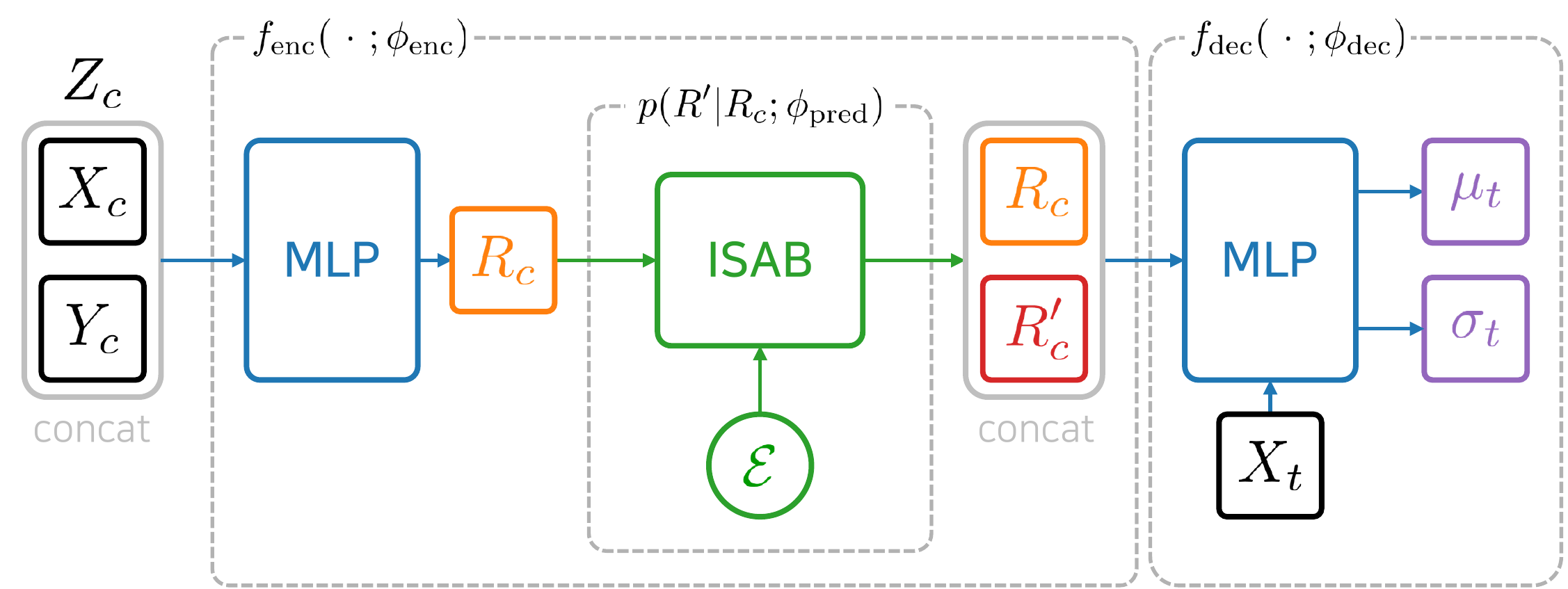}
    \caption{Concept figure of our feature generating model applied to \gls{cnp}~\citep{garnelo2018conditional}. We first convert given context dataset $Z_c$ to the representation $R_c$ using \gls{mlp} layers. Next we sample $\epsilon$ from a simple distribution (e.g. Gaussian). Then we generate the pseudo context representation $R_c'$ using generator as one layer \gls{isab}~\citep{lee2019set} in our experiment.
    }
    \label{figure/main_concept_mpnp}
\end{figure}
When $z$ is low-dimensional, it would be moderately easy to learn the joint predictives, but in practice, we often encounter problems with high-dimensional $z$, for instance when the input $x$ is a high-resolution image. For such cases, directly generating $z$ may be harder than the original problem, severely deteriorating the overall learning procedure of \gls{mpnp}.
Instead, we propose to generate the \emph{encoded representations} of $z$. The encoders of the most of the \glspl{npf} first encode an input $z_i$ into a representation $r_i$. For the remaining of the forward pass, we only need $r_i$s instead of the original input $z$. Hence we can build a joint predictives $p(R'|R_c; \phi_\text{pred})$ generating $R' = \{r_i'\}_{i=1}^{N-|c|}$ conditioned on $R_c = \{r_i\}_{i\in c}$ as for generating $Z'$ from $Z_c$. In the experiments, we compare these two versions of \glspl{mpnp} (generating $Z'$ and generating $R'$), and found that the one generating $R'$ works much better both in terms of data efficiency in training and predictive performances, even when the dimension of $z$ is not particularly large.
See \cref{figure/main_concept_mpnp} for our method applying to \gls{cnp} model~\citep{garnelo2018conditional}.

\subsection{Training}\label{main:subsec:training}

With the generator $p(Z'|Z_c;\phi_\text{pred})$, the marginal likelihood for a task $\tau = (Z, c)$ is computed as
\[\label{eq:logmarginal_mp}
\log p(Y|X, Z_c) = \log\int \exp\bigg(-\sum_{i\in [n]}\ell(z_i, \tilde\theta(Z_c\cup Z'))\bigg) p(Z'|Z_c;\phi_\text{pred}) \dee Z'.
\]
Note that $p(Z'|Z_c;\phi_{\text{pred}})$ is \gls{cid}, so there exists a corresponding martingale posterior $\pi_N$ such that
\[
\log p(Y|X, Z_c) = \log\int \exp\bigg(-\sum_{i\in [n]}\ell(z_i, \theta)\bigg) \pi_N(\theta|Z_c) \dee \theta.
\]
We approximate the marginal likelihood via a consistent estimator,
\[\label{eq:mpnp_term1}
\log p(Y|X, Z_c) \approx \log \Bigg[ \frac{1}{K} \sum_{k=1}^K \exp\bigg(-\sum_{i\in [n]}\ell(z_i, \tilde\theta(Z_c\cup Z'^{(k)}))\bigg)
\Bigg] := -\calL_\text{marg}(\tau,\phi),
\]
where $Z'^{(1)},\dots, Z'^{(K)}\iidsim p(Z'|Z_c;\phi_\text{pred})$. This objective would be suffice if we are given sufficiently good $\tilde\theta(Z_c \cup Z'^{(k)})$, but we have to also train the encoder to properly amortize the parameter construction process \cref{eq:recovering_theta}. For this, we use only the given context data to optimize
\[\label{eq:mpnp_term2}
\log p_{\textsc{cnp}}(Y|X, Z_c) = -\sum_{i\in [n]} \ell(z_i, \tilde\theta(Z_c)) := -\calL_\text{amort}(\tau,\phi)
\]
that is, we train the parameters $(\phi_\text{enc}, \phi_\text{dec})$ using \gls{cnp} objective. Furthermore, we found that if we just maximize \cref{eq:mpnp_term1} and \cref{eq:mpnp_term2}, the model can cheat by ignoring the generated pseudo contexts and use only the original context to build function estimates. To prevent this, we further maximize the similar \gls{cnp} objectives for each generated pseudo context to encourage the model to actually make use of the generated contexts.
\[\label{eq:mpnp_term3}
\frac{1}{K}\sum_{k=1}^K \log p_{\textsc{cnp}} (Y|X, Z'^{(k)}) = -\frac{1}{K}\sum_{i\in [n]} \ell(z_i, \tilde\theta(Z'^{(k)})) := -\calL_\text{pseudo}(\tau,\phi)
\]
Combining these, the loss function for the \gls{mpnp} is then
\[\label{eq:mpnp_term_whole}
\bbE_{\tau}[\calL(\tau,\phi)] = \bbE_\tau[\calL_\text{marg}(\tau,\phi) + \calL_\text{amort}(\tau,\phi) + \calL_\text{pseudo}(\tau,\phi)].
\]

\section{Related Works}
\label{main:sec:related}
\gls{cnp}~\citep{garnelo2018conditional} is the first \gls{npf} model which consists of simple \gls{mlp} layers as its encoder and decoder. \gls{np}~\citep{garnelo2018neural} also uses \gls{mlp} layers as its encoder and decoder but introduces a global latent variable to model a functional uncertainty.
\gls{canp}~\citep{kim2018attentive} and \gls{anp}~\citep{kim2018attentive} are the models which apply attention modules as their encoder block in order to well summarize context information relevant to target points. 
\citet{louizos2019functional} proposed \glspl{np} model which employs local latent variables instead of a global latent variable by applying a graph neural network.
By applying convolution layers as their encoder, \citet{gordon2020convolutional} and \citet{foong2020meta} introduced a translation equivariant \glspl{cnp} and \glspl{np} model, respectively. 
In addition to these works, \gls{bnp}~\citep{lee2020bootstrapping} suggests modeling functional uncertainty with the bootstrap~\citep{efron1992bootstrap} method instead of using a single global latent variable. 

\section{Experiments}
\label{main:sec:experiments}

We provide extensive experimental results to show how \gls{mpnp} and \gls{mpanp} effectively increase performance upon the following baselines: \gls{cnp}, \gls{np}, \gls{bnp}, \gls{canp}, \gls{anp}, and \gls{banp}. All models except deterministic models (i.e., \gls{cnp} and \gls{canp}) use the same number of samples; $K=5$ for the image completion task and $K=10$ for the others. Refer to~\cref{app:sec:architectures,app:sec:details} for more detailed experimental setup including model architectures, dataset and evaluation metrics.


\subsection{1D Regression}
\label{main:sec:experiments:1dregression}

In this section, we conducted 1D regression experiments following \citet{kim2018attentive} and \citet{lee2020bootstrapping}. 
In this experiments, the dataset curves are generated from \gls{gp} with 4 different settings: \romannumeral1) RBF kernels, \romannumeral2) Mat\'ern 5/2 kernels, \romannumeral3) Periodic kernels, and \romannumeral4) RBF kernels with Student's $t$ noise.

\paragraph{Infinite Training Dataset}
\label{main:subsec:infinite_training}
\begin{figure}[t]
    \centering
    \includegraphics[width = 0.49\textwidth]{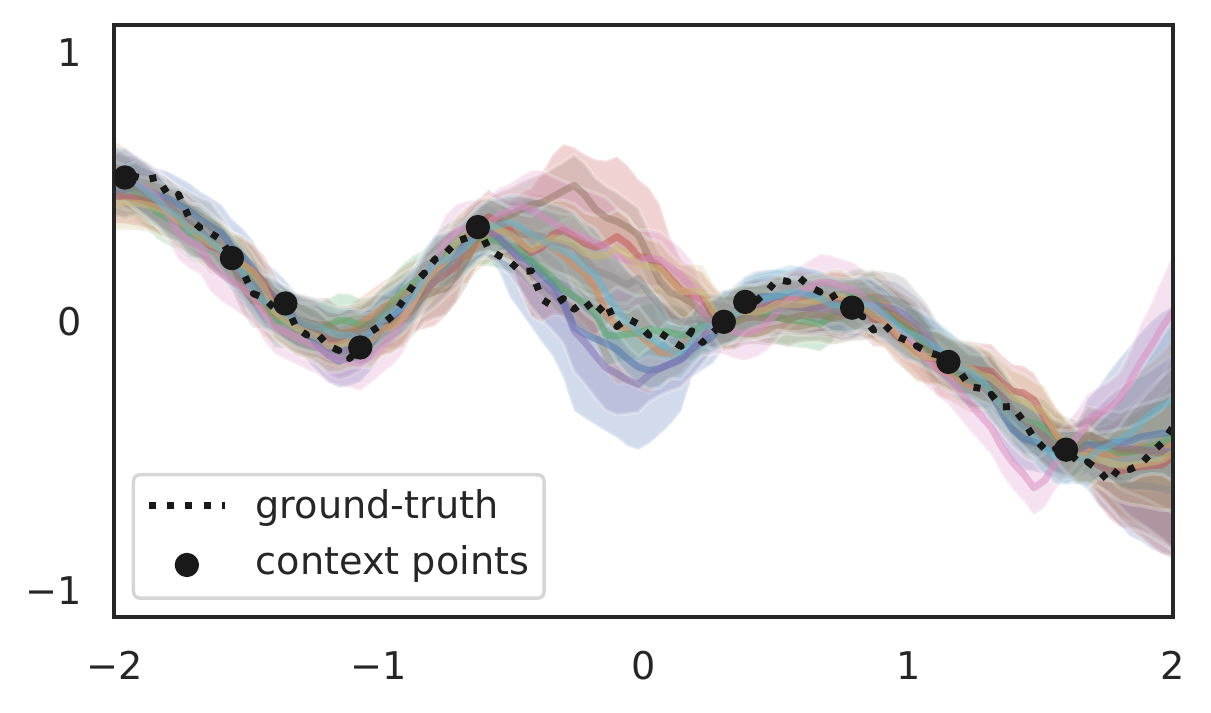}
    \includegraphics[width = 0.49\textwidth]{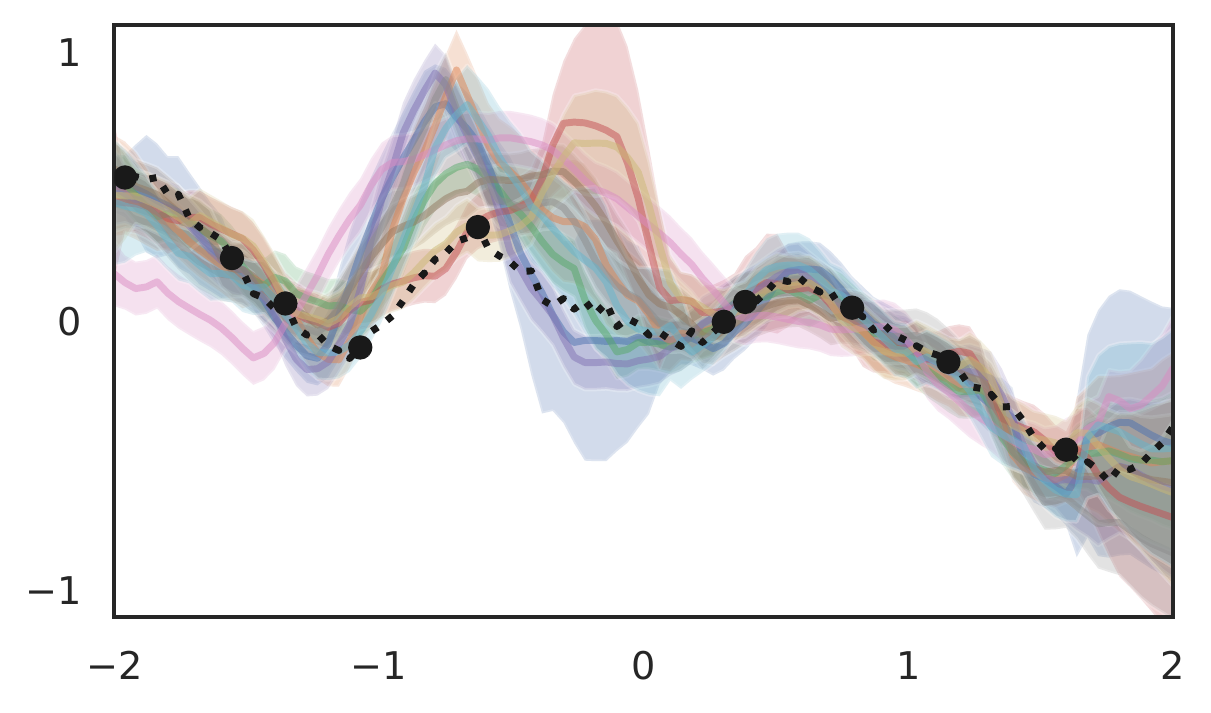}
    \caption{Posterior samples of \gls{mpanp} for 1D regression task with RBF kernel. The black dashed line is the true function sampled from \gls{gp} with RBF kernel, and the black dots are context points. We visualized decoded mean and standard deviation with colored lines and areas. (Left) \Gls{mpanp} posterior predictions using the combined features of the original contexts and the generated pseudo contexts. (Right) Predictions using only the generated pseudo contexts without the original contexts. The pseudo contexts are decoded into reasonable functions, especially with high uncertainty for the region without context points.}
    \label{fig:feature_method}
\end{figure}

\begin{table}[t]
\centering
\caption{Test results for 1D regression tasks on RBF, Matern, Periodic, and $t$-noise. `Context' and `Target' respectively denote context and target log-likelihood values. All values are averaged over four seeds. See~\cref{table/app_gp_inf_full} for the task log-likelihood values.}
\label{table/main_gp_inf}
\resizebox{\linewidth}{!}{
\begin{tabular}{lrrrrrrrr}
\toprule
      & \multicolumn{2}{c}{RBF} & \multicolumn{2}{c}{Matern} & \multicolumn{2}{c}{Periodic} & \multicolumn{2}{c}{$t$-noise} \\
\cmidrule(lr){2-3}\cmidrule(lr){4-5}\cmidrule(lr){6-7}\cmidrule(lr){8-9}
Model & Context & Target & Context & Target & Context & Target & Context & Target \\
\midrule
CNP & 
 1.096$\spm{0.023}$ &  0.515$\spm{0.018}$ &
 1.031$\spm{0.010}$ &  0.347$\spm{0.006}$ &
-0.120$\spm{0.020}$ & -0.729$\spm{0.004}$ &
 0.032$\spm{0.014}$ & -0.816$\spm{0.032}$ \\
NP & 
 1.022$\spm{0.005}$ &  0.498$\spm{0.003}$ &
 0.948$\spm{0.006}$ &  0.337$\spm{0.005}$ &
-0.267$\spm{0.024}$ & \textBF{-0.668}$\spm{0.006}$ &
 \textBF{0.201}$\spm{0.025}$ & -0.333$\spm{0.078}$ \\
BNP & 
 1.112$\spm{0.003}$ &  0.588$\spm{0.004}$ &
 1.057$\spm{0.009}$ &  0.418$\spm{0.006}$ &
-0.106$\spm{0.017}$ & -0.705$\spm{0.001}$ &
-0.009$\spm{0.032}$ & -0.619$\spm{0.191}$ \\
\textBF{MPNP (ours)} & 
 \textBF{1.189}$\spm{0.005}$ &  \textBF{0.675}$\spm{0.003}$ &
 \textBF{1.123}$\spm{0.005}$ &  \textBF{0.481}$\spm{0.007}$ &
 \textBF{0.205}$\spm{0.020}$ & \textBF{-0.668}$\spm{0.008}$ &
 0.145$\spm{0.017}$ & \textBF{-0.329}$\spm{0.025}$ \\
\midrule
CANP & 
 1.304$\spm{0.027}$ &  0.847$\spm{0.005}$ &
 1.264$\spm{0.041}$ &  0.662$\spm{0.013}$ &
 0.527$\spm{0.106}$ & -0.592$\spm{0.002}$ &
 0.410$\spm{0.155}$ & -0.577$\spm{0.022}$ \\
ANP & 
 \textBF{1.380}$\spm{0.000}$ &  0.850$\spm{0.007}$ &
 \textBF{1.380}$\spm{0.000}$ &  0.663$\spm{0.004}$ &
 0.583$\spm{0.011}$ & -1.019$\spm{0.023}$ &
 0.836$\spm{0.071}$ & -0.415$\spm{0.131}$ \\
BANP & 
 \textBF{1.380}$\spm{0.000}$ &  0.846$\spm{0.001}$ &
 \textBF{1.380}$\spm{0.000}$ &  0.662$\spm{0.005}$ &
 \textBF{1.354}$\spm{0.006}$ & -0.496$\spm{0.005}$ &
 0.646$\spm{0.042}$ & -0.425$\spm{0.050}$ \\
\textBF{MPANP (ours)} & 
 1.379$\spm{0.000}$ &  \textBF{0.881}$\spm{0.003}$ &
 \textBF{1.380}$\spm{0.000}$ &  \textBF{0.692}$\spm{0.003}$ &
 1.348$\spm{0.005}$ & \textBF{-0.494}$\spm{0.007}$ &
 \textBF{0.842}$\spm{0.062}$ & \textBF{-0.332}$\spm{0.026}$ \\
\bottomrule
\end{tabular}}
\end{table}

\begin{figure}[t]
    \centering
    \includegraphics[width=0.95\textwidth]{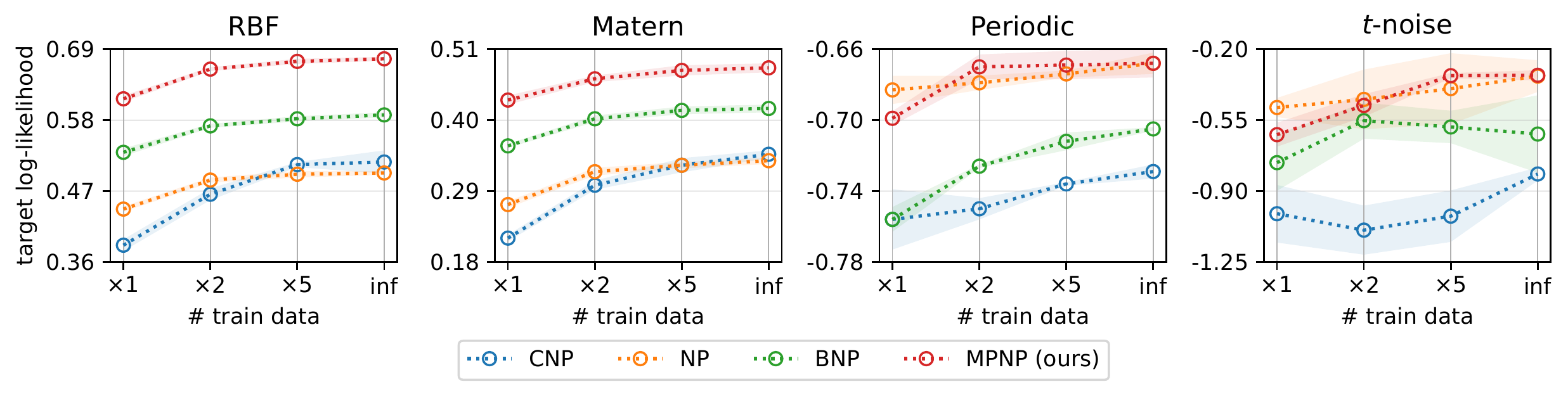}
    \includegraphics[width=0.95\textwidth]{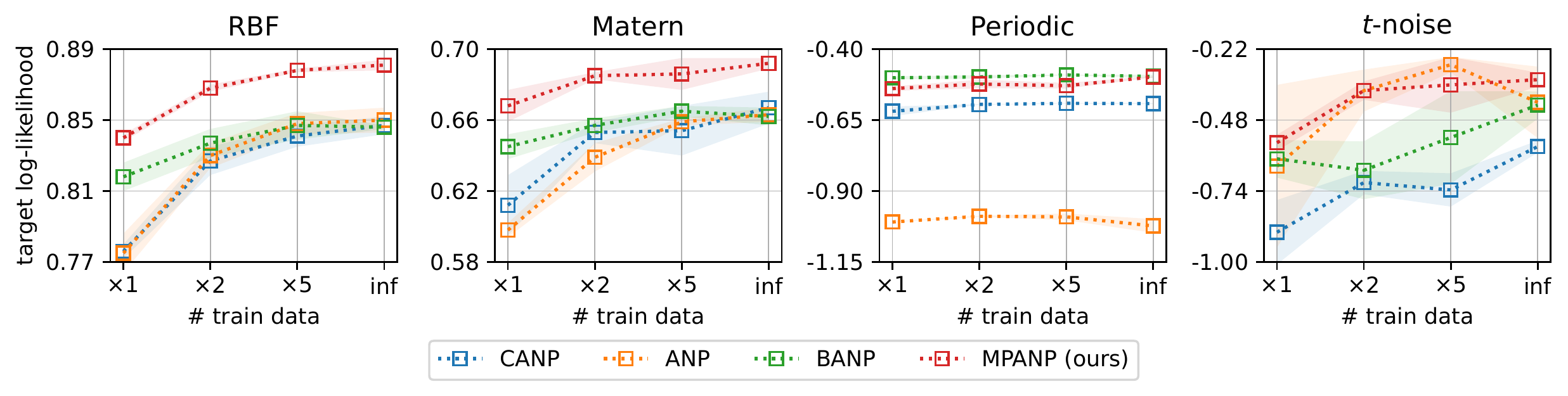}
    \caption{Test target log-likelihood values with varying the number of train data for 1D regression tasks on RBF, Matern, Periodic, and $t$-noise. Here, x-axis denotes how many examples are used for training, i.e., $\times1$, $\times2$, and $\times5$ respectively denote $51200$, $102400$, and $256000$ train examples.}
    \label{figure/main_gp_finite}
\end{figure}

Previous works~\citep{garnelo2018neural,kim2018attentive,le2018empirical} assumed that there exists a \gls{gp} curve generator that can provide virtually infinite amount of tasks for training. We first follow this setup, training all models for 100,000 steps where a new task is generated from each training step. We compare the models by picking checkpoints achieving the lowest validation loss. \cref{table/main_gp_inf} clearly shows that our model outperforms the other models in most cases. This results show that our model well captures the functional uncertainty compared to the other methods.
In~\cref{app:sec:additional_experiments}, we also report the comparison with the baselines with increased number of parameters to match the additional number of parameters introduced for the generator in our model, where ours still significantly outperforms the baselines.


\paragraph{Finite Training Dataset}
We also compare the models on more realistic setting assuming a finite amount of training tasks. Specifically, we first configured the finite training dataset consisting of $\{51200, 102400, 256000\}$ examples at the start of the training, instead of generating new tasks for each training step. We then trained all models with the same 100,000 training iterations in order to train the models with the same training budget as in the infinite training dataset situation. \cref{figure/main_gp_finite} clearly shows that our model consistently outperforms other models in terms of the target log-likelihood even when the training dataset is finite. This indicates that \glspl{mpnp} effectively learn a predictive distribution of unseen dataset from a given dataset with small number of tasks. Refer to~\cref{app:sec:additional_experiments} for more detailed results.

\subsection{Image Completion}
\label{main:sec:experiments:imagecompletion}

\begin{table}[t]
\centering
\caption{Test results for image completion tasks on MNIST, SVHN, and CelebA. `Context' and `Target' respectively denote context and target log-likelihood values, and `Task' denotes the task log-likelihood. All values are averaged over four seeds.}
\label{table/main_image}
\resizebox{\linewidth}{!}{
\begin{tabular}{lrrrrrrrrr}
\toprule
      & \multicolumn{3}{c}{MNIST} & \multicolumn{3}{c}{SVHN} & \multicolumn{3}{c}{CelebA} \\
\cmidrule(lr){2-4}\cmidrule(lr){5-7}\cmidrule(lr){8-10}
Model & Context & Target & Task & Context & Target & Task & Context & Target & Task \\
\midrule
CNP                   &
0.878$\spm{0.016}$ & 0.690$\spm{0.010}$ & 0.706$\spm{0.011}$ &
3.009$\spm{0.069}$ & 2.785$\spm{0.053}$ & 2.796$\spm{0.054}$ &
2.692$\spm{0.018}$ & 2.099$\spm{0.011}$ & 2.134$\spm{0.012}$ \\
NP                    &
0.797$\spm{0.004}$ & 0.707$\spm{0.004}$ & 0.714$\spm{0.003}$ &
3.045$\spm{0.021}$ & 2.841$\spm{0.019}$ & 2.851$\spm{0.019}$ &
2.721$\spm{0.017}$ & 2.216$\spm{0.013}$ & 2.246$\spm{0.013}$ \\
BNP                   &
0.859$\spm{0.050}$ & 0.742$\spm{0.026}$ & 0.752$\spm{0.029}$ &
3.169$\spm{0.028}$ & 2.946$\spm{0.023}$ & 2.957$\spm{0.023}$ &
2.897$\spm{0.011}$ & 2.329$\spm{0.010}$ & 2.394$\spm{0.010}$ \\
\textBF{MPNP (ours)}  &
\textBF{0.861}$\spm{0.010}$ & \textBF{0.747}$\spm{0.005}$ & \textBF{0.757}$\spm{0.005}$ &
\textBF{3.220}$\spm{0.017}$ & \textBF{2.980}$\spm{0.016}$ & \textBF{2.992}$\spm{0.016}$ &
\textBF{2.997}$\spm{0.010}$ & \textBF{2.369}$\spm{0.006}$ & \textBF{2.407}$\spm{0.006}$ \\
\midrule
CANP                  &
0.871$\spm{0.020}$ & 0.688$\spm{0.012}$ & 0.685$\spm{0.013}$ &
3.079$\spm{0.052}$ & 3.386$\spm{0.020}$ & 3.335$\spm{0.023}$ &
2.695$\spm{0.033}$ & 2.674$\spm{0.011}$ & 2.642$\spm{0.011}$ \\
ANP                   &
1.186$\spm{0.050}$ & 0.744$\spm{0.008}$ & 0.793$\spm{0.009}$ &
3.996$\spm{0.064}$ & 3.365$\spm{0.053}$ & 3.405$\spm{0.053}$ &
4.086$\spm{0.024}$ & 2.724$\spm{0.029}$ & 2.833$\spm{0.026}$ \\
BANP                  &
1.329$\spm{0.021}$ & 0.752$\spm{0.018}$ & 0.819$\spm{0.018}$ &
4.019$\spm{0.017}$ & 3.437$\spm{0.026}$ & 3.476$\spm{0.024}$ &
4.126$\spm{0.003}$ & 2.764$\spm{0.020}$ & 2.871$\spm{0.018}$ \\
\textBF{MPANP (ours)} &
\textBF{1.361}$\spm{0.008}$ & \textBF{0.798}$\spm{0.003}$ & \textBF{0.862}$\spm{0.003}$ &
\textBF{4.117}$\spm{0.003}$ & \textBF{3.502}$\spm{0.026}$ & \textBF{3.544}$\spm{0.024}$ &
\textBF{4.136}$\spm{0.001}$ & \textBF{2.833}$\spm{0.010}$ & \textBF{2.934}$\spm{0.009}$ \\
\bottomrule
\end{tabular}}
\end{table}

Next we conducted 2D image completion tasks for three different datasets, i.e., MNIST, SVHN, and CelebA.
For training, we uniformly sample the number of context pixels $|c|\in\{3,...,197\}$ and the number of target pixels $|t|\in\{3,...,200-|c|\}$ from an image. For evaluation, we uniformly sample the number of context pixels $|c|\in\{3,...,197\}$ and set all the remaining pixels as the targets. \cref{table/main_image} clearly demonstrates that our model outperforms the baselines over all three datasets, demonstrating the effectiveness of our method for high-dimensional image data. See~\cref{app:sec:additional_experiments} for the visualizations of completed images along with the uncertainties in terms of predictive variances, and~\cref{app:sec:details} for the detailed training setup.

\subsection{Bayesian Optimization}
\label{main:sec:experiments:bo}
\begin{figure}
\centering
\includegraphics[width=0.49\linewidth]{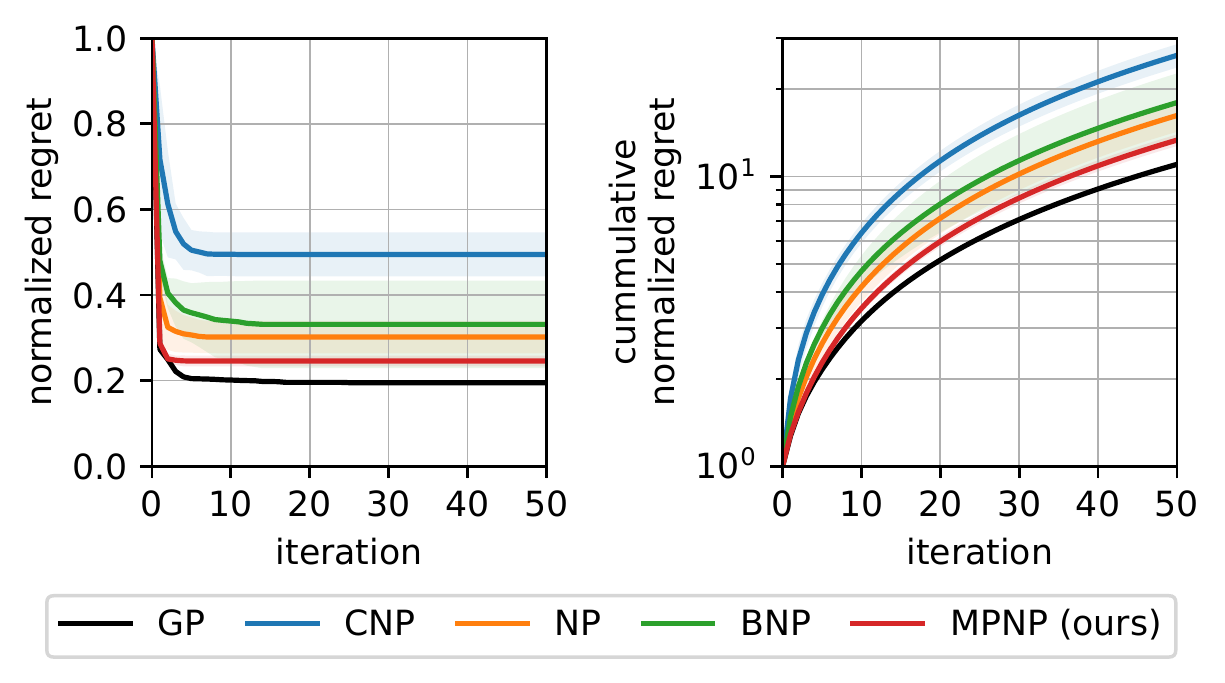}
\includegraphics[width=0.49\linewidth]{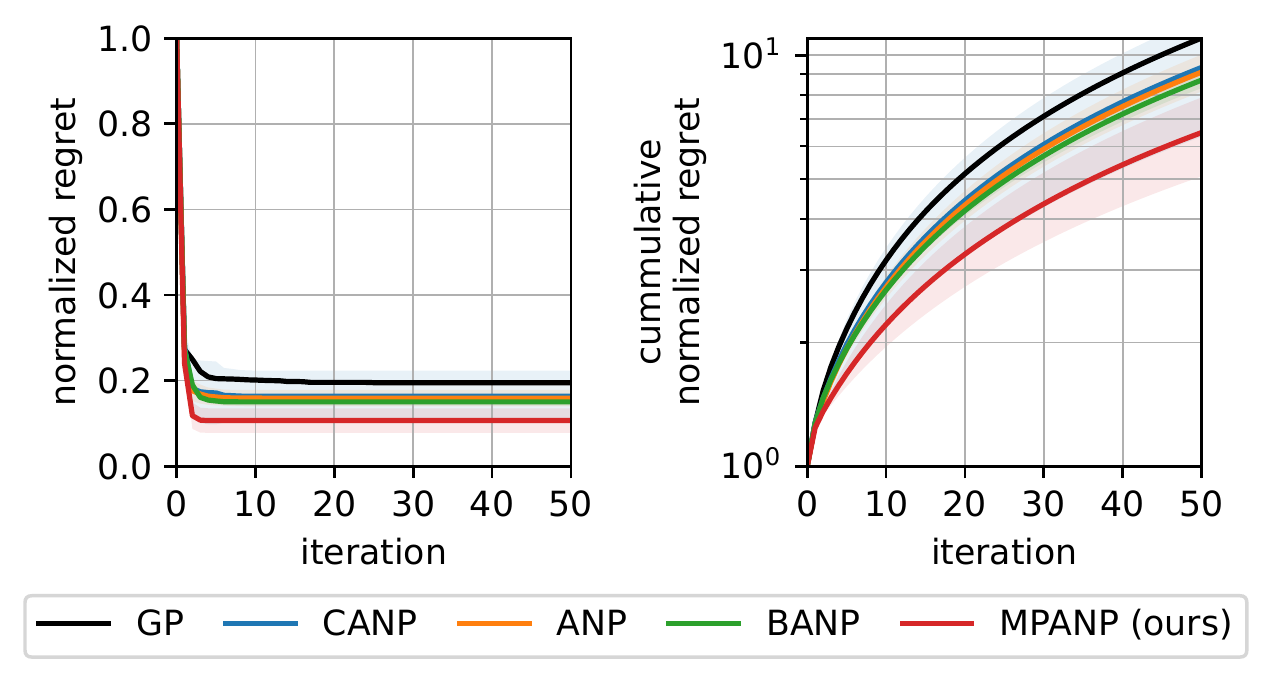}
\caption{Results for Bayesian optimization on \citet{gramacy2012cases} function; we measured normalized simple regret and its cumulative value for a iteration. All models are pre-trained on 1D regression task generated with RBF kernel (cf.~\cref{main:sec:experiments:1dregression}) and evaluated on the benchmark function for Bayesian optimization.}
\label{figure/main_bo_gram}
\end{figure}

Using pre-trained models with RBF kernels in~\cref{main:subsec:infinite_training} Infinite Training Dataset experiments, we conducted Bayesian optimization~\citep{brochu2010tutorial} for two benchmark functions~\citep{gramacy2012cases,sobester2008engineering}. As a performance measurement, we use best simple regret, which measures the difference between the current best value and the global optimum value. \cref{figure/main_bo_gram} depicts the normalized regret and the cumulative normalized regret averaged over 100 trials of the~\citet{gramacy2012cases} function.
Here, we also consider a \gls{gp} variant with RBF kernel, tuned by pre-training~\citep{wang2021automatic}.
It clearly demonstrates that our model shows the best performance among \glspl{np} for both the normalized regret and the cumulative normalized regret. \cref{app:sec:additional_experiments:bo} provides the results for the~\citet{sobester2008engineering} function and \cref{app:sec:details:bo} provides detailed experimental setups.

\subsection{Predator-Prey Model}
\label{main:subsec:predator-prey}
Following~\citet{lee2020bootstrapping}, we conducted the predator-prey population regression experiments. We first trained the models using the simulation datasets which are generated from a Lotka-Volterra model~\citep{wilkinson2018stochastic} with the simulation settings followed by~\citet{lee2020bootstrapping}. Then tested on the generated simulation test dataset and real-world dataset which is called Hudson's Bay hare-lynx data. As mentioned in~\citet{lee2020bootstrapping}, the real-world dataset shows different tendency from generated simulation datasets, so we can treat this experiment as model-data mismatch experiments. In~\cref{table/main_lv}, we can see the \glspl{mpnp} outperform the other baselines for the test simulation datasets but underperforms in the real-world dataset compare to other baselines. This shows that model-data mismatch is an open problem for the \glspl{mpnp}.

\begin{table}[t]
\centering
\caption{Test results for predator-prey population regression tasks on Lotka-Volterra simulated data and real data. `Context' and `Target' respectively denote context and target log-likelihood values, and `Task' denotes the task log-likelihood. All values are averaged over four seeds.}
\label{table/main_lv}
\resizebox{0.8\linewidth}{!}{
\begin{tabular}{lrrrrrr}
\toprule
      & \multicolumn{3}{c}{Simulated data} & \multicolumn{3}{c}{Real data} \\
\cmidrule(lr){2-4}\cmidrule(lr){5-7}
Model & Context & Target & Task & Context & Target & Task \\
\midrule
CNP & 
 0.327$\spm{0.036}$ &  0.035$\spm{0.029}$ &  0.181$\spm{0.032}$ &
-2.686$\spm{0.024}$ & -3.201$\spm{0.042}$ & -3.000$\spm{0.034}$ \\
NP & 
 0.112$\spm{0.063}$ & -0.115$\spm{0.057}$ &  0.000$\spm{0.060}$ &
-2.770$\spm{0.028}$ & -3.144$\spm{0.031}$ & -2.993$\spm{0.029}$ \\
BNP & 
 0.550$\spm{0.057}$ &  0.274$\spm{0.042}$ &  0.417$\spm{0.050}$ &
\textBF{-2.614}$\spm{0.050}$ & \textBF{-3.052}$\spm{0.022}$ & \textBF{-2.868}$\spm{0.024}$ \\
\textBF{MPNP (ours)}  & 
 \textBF{0.626}$\spm{0.041}$ &  \textBF{0.375}$\spm{0.036}$ &  \textBF{0.500}$\spm{0.038}$ &
-2.621$\spm{0.072}$ & -3.092$\spm{0.054}$ & -2.918$\spm{0.061}$ \\
\midrule
CANP & 
 0.689$\spm{0.046}$ &  1.615$\spm{0.026}$ &  1.023$\spm{0.018}$ &
-4.743$\spm{1.119}$ & -6.413$\spm{0.339}$ & -5.801$\spm{0.733}$ \\
ANP & 
 2.607$\spm{0.015}$ &  1.830$\spm{0.020}$ &  2.234$\spm{0.018}$ &
 1.887$\spm{0.078}$ & -4.848$\spm{0.385}$ & -1.615$\spm{0.188}$ \\
BANP & 
 \textBF{2.654}$\spm{0.000}$ &  1.797$\spm{0.012}$ &  2.240$\spm{0.006}$ &
 \textBF{2.190}$\spm{0.062}$ & \textBF{-3.597}$\spm{0.279}$ & \textBF{-0.741}$\spm{0.160}$ \\
\textBF{MPANP (ours)} & 
 2.639$\spm{0.008}$ &  \textBF{1.835}$\spm{0.004}$ &  \textBF{2.254}$\spm{0.006}$ &
 1.995$\spm{0.145}$ & -5.073$\spm{0.680}$ & -1.690$\spm{0.401}$ \\
\bottomrule
\end{tabular}}
\end{table}

\section{Conclusion}
\label{main:sec:conclusion}
In this paper, we proposed a novel extension of \glspl{np} by taking a new approach to model the functional uncertainty for \glspl{np}.
The proposed model \gls{mpnp} utilizes the martingale posterior distribution~\citep{fong2021martingale}, 
where the functional uncertainty is driven from the uncertainty of future data generated from the joint predictive.
We present a simple architecture satisfying the theoretical requirements of the martingale posterior, and propose a training scheme to properly train it. We empirically validate \glspl{mpnp} on various tasks, where our method consistently outperforms the baselines.

\paragraph{Limitation}
As we presented in the \textbf{Predator-Prey Model} experiments in \cref{main:subsec:predator-prey}, 
our method did not significantly outperform baselines under model-data mismatch. This was also higlighted in \citet{fong2021martingale}: model-data mismatch under the martingale posterior framework remains an open problem.
Our method with direct input generation also performed poorly, as we found it difficult to prevent models from generating meaningless inputs that are ignored by the decoders. We present more details on unsuccessful attempts for direct input generation in \cref{app:sec:directly_generating_input}.

\clearpage
\newpage
\paragraph{Societal Impacts}
Our work is unlikely to bring any negative societal impacts. Modeling functional uncertainty may be related to the discussion of safe AI within the community.

\paragraph{Reproducibility Statement}
We argued our experimental details in \cref{app:sec:details} which contains used libraries and hardwares.
We presented all the dataset description in \cref{app:sec:details}.
We describes the model architecture details in \cref{app:sec:architectures}.

\section*{Acknowledgements}
This work was partly supported by Institute of Information \& communications Technology Planning \& Evaluation (IITP) grant funded by the Korea government(MSIT) 
(No.2019-0-00075, Artificial Intelligence Graduate School Program(KAIST)), Institute of Information \& communications Technology Planning \& Evaluation (IITP) grant funded by the Korea government(MSIT) (No.2022-0-00713), and Institute of Information \& communications Technology Planning \& Evaluation (IITP) grant funded by the Korea government(MSIT) (No.2021-0-02068, Artificial Intelligence Innovation Hub).
\bibliography{references}

\clearpage
\newpage
\appendix

\section{Model Architectures}
\label{app:sec:architectures}

\begin{figure}[t]
    \centering
    \includegraphics[width = \linewidth]{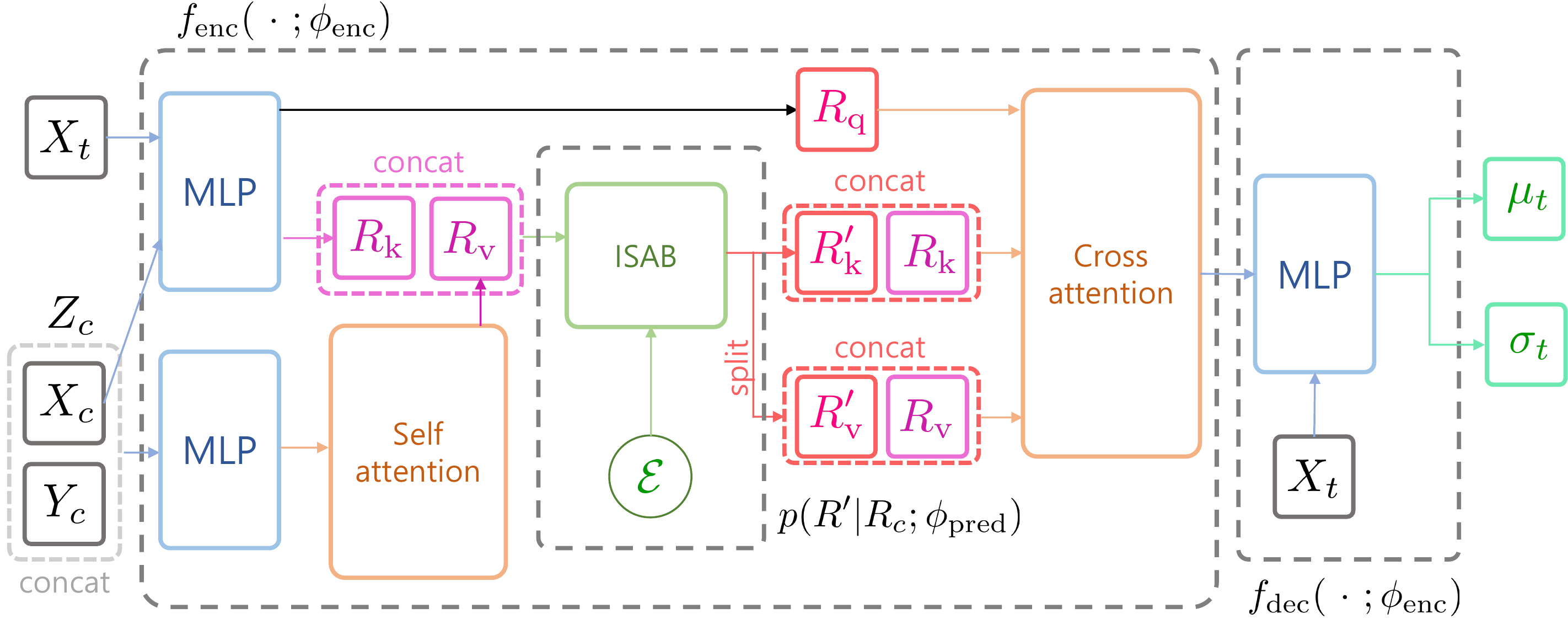}
    \caption{Concept figure of our feature generating model applied to \gls{canp}~\citep{kim2018attentive}. Here we sample $\epsilon$ from a simple distribution (e.g. Gaussian). We generate key feature $R_k'$ and value feature $R_v'$ for cross attention layer which are corresponding to pseudo context data. We use generator as one layer \gls{isab}~\citep{lee2019set} in our experiment.
    }
    \label{figure/main_concept_mpanp}
\end{figure}

In this section, we summarize the model architectures which we used in experiments.
Here, we only present simplified structures for each model.
To see exact computation procedures for \glspl{bnp}, please refer to \citet{lee2020bootstrapping}. \cref{figure/main_concept_mpanp} shows our method applying to \gls{canp} model~\citep{kim2018attentive}.

\subsection{Modules}

\paragraph{Linear Layer} $\Lin(d_{\In},d_{\out})$ denotes the linear transformation of the input with dimension $d_{\In}$ into the output with dimension $d_{\out}$.
\paragraph{Multi-Layer Perceptron} $\MLP(n_l, d_{\In}, d_{\hid}, d_{\out})$ denotes a multi-layer perceptron with the structure:
\begin{align*}
    \MLP(n_l, d_{\In}, d_{\hid}, d_{\out}) = \Lin(d_{\hid}, d_{\out})\circ(\ReLU\circ\Lin(d_{\hid},d_{\hid}))^{n_l-2}\circ\ReLU\circ\Lin(d_{\In},d_{\hid}),
\end{align*}
where $\ReLU$ denotes the element-wise Rectified Linear Unit (ReLU) activation function.
\paragraph{Multi-Head Attention} $\MHA(n_{\head}, d_{\out})(Q,K,V)$ denotes a multi-head attention~\citep{vaswani2017attention} with $n_{\head}$ heads which takes input as $(Q,K,V)$ and outputs the feature with dimension $d_{\out}$.
The actual computation of $\MHA(n_{\head}, d_{\out})(Q,K,V)$ can be written as follows:
\begin{align*}
    (Q_i')_{i=1}^{n_\head} &= \spl(\Lin(d_q,d_\out)(Q), n_\head)\\
    (K_i')_{i=1}^{n_\head} &= \spl(\Lin(d_k,d_\out)(K), n_\head)\\
    (V_i')_{i=1}^{n_\head} &= \spl(\Lin(d_v,d_\out)(V), n_\head)\\
    H &= \concat([\softmax(Q_i'K_i'^\top/\sqrt{d_\out})V_i']_{i=1}^{n_\head})\\
    O &= \LN(Q'+H)\\
    \MHA(n_\head,d_\out)(Q,K,V) &= \LN(O+\ReLU(\Lin(d_\out,d_\out)(O)))
\end{align*}
where $(d_q, d_k, d_v)$ denotes the dimension of $Q,K,V$ respectively, $\spl$ and $\concat$ are the splitting and concatenating $A$ in the feature dimension respectively, and $\LN$ denotes the layer normalization~\citep{ba2016layer}.
\paragraph{Self-Attention} $\SA(n_\head, d_\out)$ denotes a self-attention module which is simply computed as $\SA(n_\head, d_\out)(X) = \MHA(n_\head, d_\out)(X,X,X)$.
\paragraph{Multi-head Attention Block} $\MAB(n_\head, d_\out)$ denotes a multi-head attention block module~\citep{lee2019set} which is simply computed as $\MAB(n_\head, d_\out)(X,Y) = \MHA(n_\head, d_\out)(X,Y,Y)$.
\paragraph{Induced Set Attention Block} $\ISAB(n_\head, d_\out)$ denotes a induced set attention block~\citep{lee2019set} which constructed with two stacked $\MAB$ layers.
The actual computation of $\ISAB(n_\head, d_\out)(X,Y)$ can be written as follows:
\begin{align*}
    H &= \MAB(n_\head, d_\out)(Y, X)\\
    \ISAB(n_\head, d_\out)(X,Y) &= \MAB(n_\head, d_\out)(X, H).
\end{align*}

\subsection{\texorpdfstring{\gls{cnp}, \gls{np}, \gls{bnp}, \gls{neubnp} and \gls{mpnp}}{CNP, NP, BNP, NeuBNP, and MPNP}}

\paragraph{Encoder}
The models only with a deterministic encoder (\gls{cnp}, \gls{bnp}, \gls{mpnp}) use the following structure:
\begin{align*}
    r_c &= \frac{1}{|c|} \sum_{i \in c} \MLP(n_l=5, d_{\In}=d_z, d_{\hid}=128, d_{\out}=128)(z_i), \\
    f_\text{enc}(Z_c) &= r_c.
\end{align*}
For the \gls{mpnp}, $f_\enc(Z_c)$ changes into $\concat([r_c,r_c'])$ where $r_c'$ is the feature of the pseudo context data generated from generator in paragraph \textbf{Generator}.
The model also with a latent encoder (\gls{np}) uses:
\begin{align*}
    r_c &= \frac{1}{|c|} \sum_{i \in c} \MLP(n_l=5, d_{\In}=d_z, d_{\hid}=128, d_{\out}=128)(z_i), \\
    (m_c, \log s_c) &= \frac{1}{|c|} \sum_{i \in c} \MLP(n_l=2, d_{\In}=d_z, d_{\hid}=128, d_{\out}=128 \times 2)(z_i), \\
    s_c &= 0.1 + 0.9 \cdot \text{softplus}(\log s_c), \\
    h_c &= \calN(m_c, s^2_c I_h), \\
    f_\text{enc}(Z_c) &= [r_c; h_c],
\end{align*}
where $d_z = d_x + d_y$ denotes the data dimension.
Data dimensions vary through tasks, $d_x=1, d_y=1$ for 1D regression tasks, $d_x=2, d_y=1$ for MNIST image completion task,  $d_x=2, d_y=3$ for SVHN and CelebA image completion tasks, and $d_x=1, d_y=2$ for Lotka Volterra task.

\paragraph{Adaptation Layer}
\gls{bnp} uses additional adaptation layer to combine bootstrapped representation and the base representation. This can be done with a simple linear layer
\[
    \tilde{r}_c = \Lin(d_{\hid}=128, d_{\hid}=d_x+128)(\tilde{r}_c^{(pre)}).
\]

\paragraph{Decoder}
All models use a single MLP as a decoder.
The models except \gls{np} uses the following structure:
\begin{align*}
    (\mu, \log \sigma) &= \MLP(n_l=3, d_{\In}=d_x + 128, d_{\hid}=128, d_{\out}=2)(\concat([x,r_c])) \\
    \sigma &= 0.1 + 0.9 \cdot \text{softplus}(\log \sigma), \\
    f_\text{dec}(x, r_c) &= (\mu, \sigma),
\end{align*}
and \gls{np} uses:
\begin{align*}
    (\mu, \log \sigma) &= \MLP(n_l=3, d_{\In}=d_x + 128 \times 2, d_{\hid}=128, d_{\out}=2)(\concat([x,r_c,h_c])) \\
    \sigma &= 0.1 + 0.9 \cdot \text{softplus}(\log \sigma), \\
    f_\text{dec}(x, r_c, h_c) &= (\mu, \sigma).
\end{align*}

\paragraph{Generator}
\gls{mpnp} use a single $\ISAB$ module as a generator. The $\ISAB$ uses the following structure: 
\begin{align*}
    \epsilon &= \concat([\epsilon_i]_{i=1}^{n_\gen})\\
    r_c' &= \ISAB(n_\head=8, d_\out=128)(\epsilon,r_c)\\
    f_\gen(r_c) &= r_c'
\end{align*}
where $\epsilon_i$s are i.i.d. sampled from Gaussian distribution with dimension 128 and $n_\gen$ denotes a number of pseudo context data.

\subsection{\texorpdfstring{\gls{canp}, \gls{anp}, \gls{banp} and \gls{mpanp}}{CANP, ANP, BANP, NeuBANP, and MPANP}}

\paragraph{Encoder}
The models only with a deterministic encoder (\gls{canp}, \gls{banp}, \gls{neubanp} and \gls{mpanp}) use the following structure:
\begin{align*}
    r_q &= \MLP(n_l=5, d_{\In}=d_x, d_{\hid}=128, d_{\out}=128)(X), \\
    r_k &= \MLP \qquad \qquad \qquad \qquad '' \qquad \qquad \qquad  \qquad \ (X_c), \\
    r_v^{(\text{pre})} &= \MLP(n_l=5, d_{\In}=d_z, d_{\hid}=128, d_{\out}=128)(X_c), \\
    r_v &= \SA(n_\head=8, d_\out=128)(r_v^{(\text{pre})}), \\
    r_c &= \MHA(n_\head=8, d_\out=128)(r_q, r_k, r_v), \\
    f_\text{enc}(Z_c) &= r_c.
\end{align*}
For the \gls{mpanp}, $f_\enc(Z_c)$ changes into 
\begin{align*}
    r_c &= \MHA(n_\head=8, d_\out=128)(r_q, \concat([r_k,r_k']), \concat([r_v,r_v'])),\\
    f_\enc &= r_c,
\end{align*} where $r_k'$ and $r_v'$ are the key and value features of the pseudo context data generated from generator in paragraph \textbf{Generator}.

\gls{anp} constructed as:
\begin{align*}
    r_q &= \MLP(n_l=5, d_{\In}=d_x, d_{\hid}=128, d_{\out}=128)(X), \\
    r_k &= \MLP \qquad \qquad \qquad \qquad '' \qquad \qquad \qquad  \qquad \ (X_c), \\
    r_v' &= \MLP(n_l=5, d_{\In}=d_z, d_{\hid}=128, d_{\out}=128)(X_c), \\
    r_v &= \SA(n_\head=8, d_\out=128)(r_v'), \\
    r_c &= \MHA(n_\head=8, d_\out=128)(r_q, r_k, r_v), \\
    h_i' &= \MLP(n_l=2, d_{\In}=d_z, d_{\hid}=128, d_{\out}=128 \times 2)(z_i), \\
    h_i &= \SA(n_\head=8, d_\out=128)(h_i'), \\
    (m_c, \log s_c) &= \frac{1}{|c|} \sum_{i \in c} h_i, \\
    s_c &= 0.1 + 0.9 \cdot \text{softplus}(\log s_c), \\
    h_c &= \calN(m_c, s^2_c I_h), \\
    f_\text{enc}(Z_c) &= [r_c; h_c].
\end{align*}
Note that $r_q$ and $r_k$ are from the same \MLP. 

\paragraph{Adaptation Layer}
Like \gls{bnp}, \gls{banp} also uses adaptation layer with same structure to combine bootstrapped representations.

\paragraph{Decoder}
All models use the same decoder structure as their non-attentive counterparts.

\paragraph{Generator}
\gls{mpanp} use a single $\ISAB$ module as a generator. The $\ISAB$ uses the following structure: 
\begin{align*}
    \epsilon &= \concat([\epsilon_i]_{i=1}^{n_\gen})\\
    (r_k',r_v') &= \ISAB(n_\head=8, d_\out=256)(\epsilon,\concat([r_k,r_v]))\\
    f_\gen(r_k,r_v) &= (r_k',r_v')
\end{align*}
where $\epsilon_i$s are i.i.d. sampled from Gaussian distribution with dimension 256 and $n_\gen$ denotes a number of pseudo context data.

\section{Additional Experiments}
\label{app:sec:additional_experiments}
\subsection{1D Regression}
\label{app:sec:additional_experiments:1dregression}

\begin{table}[t]
\centering
\setlength{\tabcolsep}{0.3em}
\caption{Test results for 1D regression tasks on RBF, Matern, Periodic, and $t$-noise. `Context' and `Target' respectively denote context and target log-likelihood values, and `Task' denotes the task log-likelihood. All values are averaged over four seeds.}
\label{table/app_gp_inf_full}
\resizebox{\linewidth}{!}{
\begin{tabular}{lrrrrrrrrrrrr}
\toprule
      & \multicolumn{3}{c}{RBF} & \multicolumn{3}{c}{Matern} & \multicolumn{3}{c}{Periodic} & \multicolumn{3}{c}{$t$-noise} \\
\cmidrule(lr){2-4}\cmidrule(lr){5-7}\cmidrule(lr){8-10}\cmidrule(lr){11-13}
Model & Context & Target & Task & Context & Target & Task & Context & Target & Task & Context & Target & Task \\
\midrule
CNP & 
 1.096$\spm{0.023}$ &  0.515$\spm{0.018}$ &  0.796$\spm{0.020}$ &
 1.031$\spm{0.010}$ &  0.347$\spm{0.006}$ &  0.693$\spm{0.008}$ &
-0.120$\spm{0.020}$ & -0.729$\spm{0.004}$ & -0.363$\spm{0.012}$ &
 0.032$\spm{0.014}$ & -0.816$\spm{0.032}$ & -0.260$\spm{0.012}$ \\
NP & 
 1.022$\spm{0.005}$ &  0.498$\spm{0.003}$ &  0.748$\spm{0.004}$ &
 0.948$\spm{0.006}$ &  0.337$\spm{0.005}$ &  0.641$\spm{0.005}$ &
-0.267$\spm{0.024}$ & \textBF{-0.668}$\spm{0.006}$ & -0.441$\spm{0.013}$ &
 \textBF{0.201}$\spm{0.025}$ & -0.333$\spm{0.078}$ & \textBF{-0.038}$\spm{0.026}$ \\
BNP & 
 1.112$\spm{0.003}$ &  0.588$\spm{0.004}$ &  0.841$\spm{0.003}$ &
 1.057$\spm{0.009}$ &  0.418$\spm{0.006}$ &  0.741$\spm{0.007}$ &
-0.106$\spm{0.017}$ & -0.705$\spm{0.001}$ & -0.347$\spm{0.010}$ &
-0.009$\spm{0.032}$ & -0.619$\spm{0.191}$ & -0.217$\spm{0.036}$ \\
\textBF{MPNP (ours)} & 
 \textBF{1.189}$\spm{0.005}$ &  \textBF{0.675}$\spm{0.003}$ &  \textBF{0.911}$\spm{0.003}$ &
 \textBF{1.123}$\spm{0.005}$ &  \textBF{0.481}$\spm{0.007}$ &  \textBF{0.796}$\spm{0.005}$ &
 \textBF{0.205}$\spm{0.020}$ & \textBF{-0.668}$\spm{0.008}$ & \textBF{-0.171}$\spm{0.013}$ &
 0.145$\spm{0.017}$ & \textBF{-0.329}$\spm{0.025}$ & -0.061$\spm{0.012}$ \\
\midrule
CANP & 
 1.304$\spm{0.027}$ &  0.847$\spm{0.005}$ &  1.036$\spm{0.020}$ &
 1.264$\spm{0.041}$ &  0.662$\spm{0.013}$ &  0.937$\spm{0.031}$ &
 0.527$\spm{0.106}$ & -0.592$\spm{0.002}$ &  0.010$\spm{0.069}$ &
 0.410$\spm{0.155}$ & -0.577$\spm{0.022}$ & -0.008$\spm{0.098}$ \\
ANP & 
 \textBF{1.380}$\spm{0.000}$ &  0.850$\spm{0.007}$ &  1.090$\spm{0.003}$ &
 \textBF{1.380}$\spm{0.000}$ &  0.663$\spm{0.004}$ &  1.019$\spm{0.002}$ &
 0.583$\spm{0.011}$ & -1.019$\spm{0.023}$ &  0.090$\spm{0.004}$ &
 0.836$\spm{0.071}$ & -0.415$\spm{0.131}$ &  0.374$\spm{0.034}$ \\
BANP & 
 \textBF{1.380}$\spm{0.000}$ &  0.846$\spm{0.001}$ &  1.088$\spm{0.000}$ &
 \textBF{1.380}$\spm{0.000}$ &  0.662$\spm{0.005}$ &  1.018$\spm{0.002}$ &
 \textBF{1.354}$\spm{0.006}$ & -0.496$\spm{0.005}$ &  \textBF{0.634}$\spm{0.005}$ &
 0.646$\spm{0.042}$ & -0.425$\spm{0.050}$ &  0.270$\spm{0.033}$ \\
\textBF{MPANP (ours)} & 
 1.379$\spm{0.000}$ &  \textBF{0.881}$\spm{0.003}$ &  \textBF{1.102}$\spm{0.001}$ &
 \textBF{1.380}$\spm{0.000}$ &  \textBF{0.692}$\spm{0.003}$ &  \textBF{1.029}$\spm{0.001}$ &
 1.348$\spm{0.005}$ & \textBF{-0.494}$\spm{0.007}$ &  0.630$\spm{0.005}$ &
 \textBF{0.842}$\spm{0.062}$ & \textBF{-0.332}$\spm{0.026}$ &  \textBF{0.384}$\spm{0.041}$ \\
\bottomrule
\end{tabular}}
\end{table}

\paragraph{Full results for~\cref{table/main_gp_inf}}
We provide the full test results for 1D regression tasks including context, target, and task log-likelihood values in~\cref{table/app_gp_inf_full}.

\paragraph{Increasing the encoder size of baselines}
\begin{table}[t]
\centering
\caption{Further comparisons with baselines with increased number of parameters. `Context' and `Target' respectively denote context and target log-liklihood values, and `Task' denotes the task log-likelihood. All values are averaged over four seeds.}
\label{table/app_gp_inf_sizeup}
\resizebox{\linewidth}{!}{
\begin{tabular}{lrrrrrrr}
\toprule
      & & \multicolumn{3}{c}{RBF} & \multicolumn{3}{c}{Matern} \\
\cmidrule(lr){3-5}\cmidrule(lr){6-8}
Model & \# Params & Context & Target & Task & Context & Target & Task \\
\midrule
CNP & 264 K & 
1.096$\spm{0.008}$ & 0.517$\spm{0.007}$ & 0.797$\spm{0.007}$ &
1.017$\spm{0.021}$ & 0.340$\spm{0.012}$ & 0.681$\spm{0.017}$ \\
NP  & 274 K & 
1.026$\spm{0.004}$ & 0.501$\spm{0.003}$ & 0.752$\spm{0.003}$ &
0.948$\spm{0.005}$ & 0.334$\spm{0.002}$ & 0.640$\spm{0.003}$ \\
BNP & 261 K & 
1.115$\spm{0.007}$ & 0.591$\spm{0.005}$ & 0.843$\spm{0.006}$ &
1.051$\spm{0.007}$ & 0.416$\spm{0.005}$ & 0.736$\spm{0.005}$ \\
\textBF{MPNP (ours)} & 266 K &
\textBF{1.189}$\spm{0.005}$ & \textBF{0.675}$\spm{0.003}$ & \textBF{0.911}$\spm{0.003}$ &
\textBF{1.123}$\spm{0.005}$ & \textBF{0.481}$\spm{0.007}$ & \textBF{0.796}$\spm{0.005}$ \\
\midrule
CANP & 868 K & 
1.305$\spm{0.007}$ & 0.844$\spm{0.006}$ & 1.035$\spm{0.005}$ &
1.278$\spm{0.013}$ & 0.663$\spm{0.006}$ & 0.947$\spm{0.008}$ \\
ANP  & 877 K & 
\textBF{1.380}$\spm{0.000}$ & 0.858$\spm{0.002}$ & 1.093$\spm{0.001}$ &
\textBF{1.380}$\spm{0.000}$ & 0.668$\spm{0.006}$ & 1.020$\spm{0.002}$ \\
BANP & 885 K & 
1.379$\spm{0.001}$ & 0.839$\spm{0.015}$ & 1.085$\spm{0.007}$ &
1.376$\spm{0.005}$ & 0.652$\spm{0.032}$ & 1.012$\spm{0.014}$ \\
\textBF{MPANP (ours)} & 877 K &
1.379$\spm{0.000}$ & \textBF{0.881}$\spm{0.003}$ & \textBF{1.102}$\spm{0.001}$ &
\textBF{1.380}$\spm{0.000}$ &  \textBF{0.692}$\spm{0.003}$ &  \textBF{1.029}$\spm{0.001}$ \\
\bottomrule
\end{tabular}}
\end{table}

Since the generator increases the size of the encoder in \glspl{mpnp}, one can claim that the performance gain of \glspl{mpnp} may come from the increased model size. To verify this, we increased the hidden dimensions of the encoder of baselines and compared them with ours. The results displayed in~\cref{table/app_gp_inf_sizeup} further clarify that ours still outperforms the baselines even when the number of parameters gets in line.

\subsection{High-D Regression}
We conducted additional experiments on the synthetic high-dimensional regression data (i.e., generating one-dimensional y from four-dimensional x with RBF kernel). Here we used the same model structures with the 1D regression task except for the input layer, and the same settings for the RBF kernel with 1D regression except for $l\sim \text{Unif}(0.5, 3.0)$. We fixed the base learning rate to $0.00015$ for all models throughout the high-dimensional regression experiments.

\cref{table/app_high_d} clearly shows our \glspl{mpnp} still outperform baselines for log-likelihood values we measured.

\begin{table}[t]
\centering

\caption{Test results for 4D regression tasks on RBF. `Context' and `Target' respectively denote context and target log-likelihood values, and `Task' denotes the task log-likelihood. All values are averaged over four seeds.}
\label{table/app_high_d}

\begin{tabular}{lrrr}
\toprule
& \multicolumn{3}{c}{RBF}  \\
\cmidrule(lr){2-4}
Model & Context & Target & Task  \\
\midrule
CNP & 0.572$\spm{0.003}$ &  0.265$\spm{0.002}$ &  0.410$\spm{0.003}$ \\
NP  & 0.568$\spm{0.009}$ &  0.267$\spm{0.004}$ &  0.407$\spm{0.007}$ \\
BNP & 0.621$\spm{0.015}$ &  0.323$\spm{0.008}$ &  0.467$\spm{0.013}$ \\
\textBF{MPNP (ours)} & 
\textBF{0.820}$\spm{0.002}$ & \textBF{0.441}$\spm{0.004}$ & \textBF{0.633}$\spm{0.004}$ \\
\midrule
CANP & 0.957$\spm{0.005}$ &  0.585$\spm{0.006}$ &  0.743$\spm{0.005}$  \\
ANP & 1.357$\spm{0.006}$ &  0.320$\spm{0.014}$ &  0.890$\spm{0.007}$  \\
BANP & \textBF{1.380}$\spm{0.000}$ &  0.549$\spm{0.006}$ &  1.013$\spm{0.002}$  \\
\textBF{MPANP (ours)} & 1.379$\spm{0.000}$ &  \textBF{0.645}$\spm{0.007}$ &  \textBF{1.046}$\spm{0.002}$ \\
\bottomrule
\end{tabular}
\end{table}

\subsection{Image Completion}
\label{app:sec:additional_experiments:imagecompletion}

\begin{figure}[t]
    \centering
    \includegraphics[width = \textwidth]{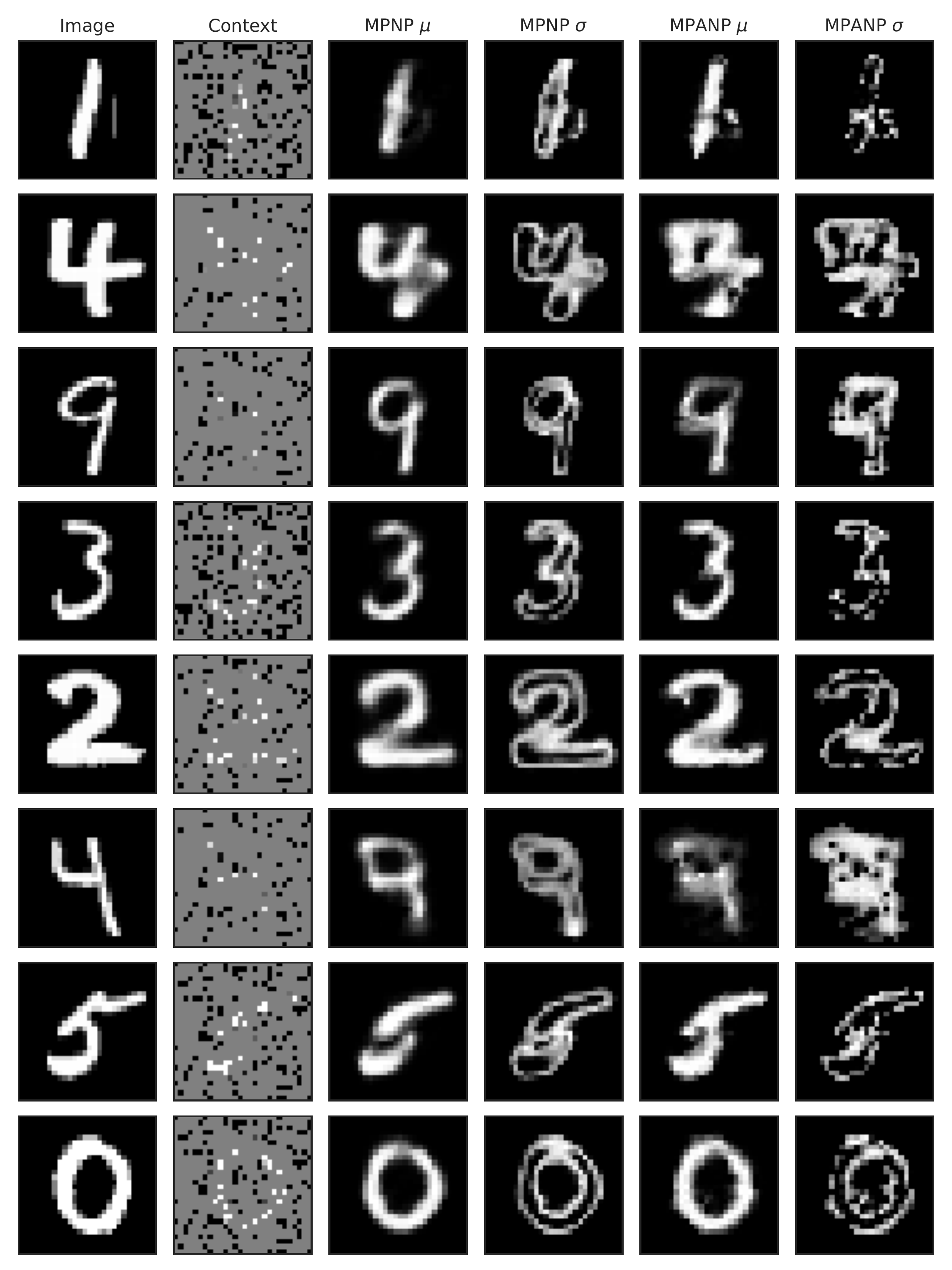}
    \caption{Predicted mean and standard deviation of image pixels by trained \glspl{mpnp} with MNIST dataset. The first column shows the real image from test dataset. The second column shows the context dataset which given to the models. The third and the forth columns show the predicted mean and standard deviation from the \gls{mpnp} respectively. The fifth and the sixth columns show the predicted mean and standard deviation from the \gls{mpanp}.} 
    \label{figure/app_visualize_mnist}
\end{figure}

\begin{figure}
    \centering
    \includegraphics[width=\linewidth]{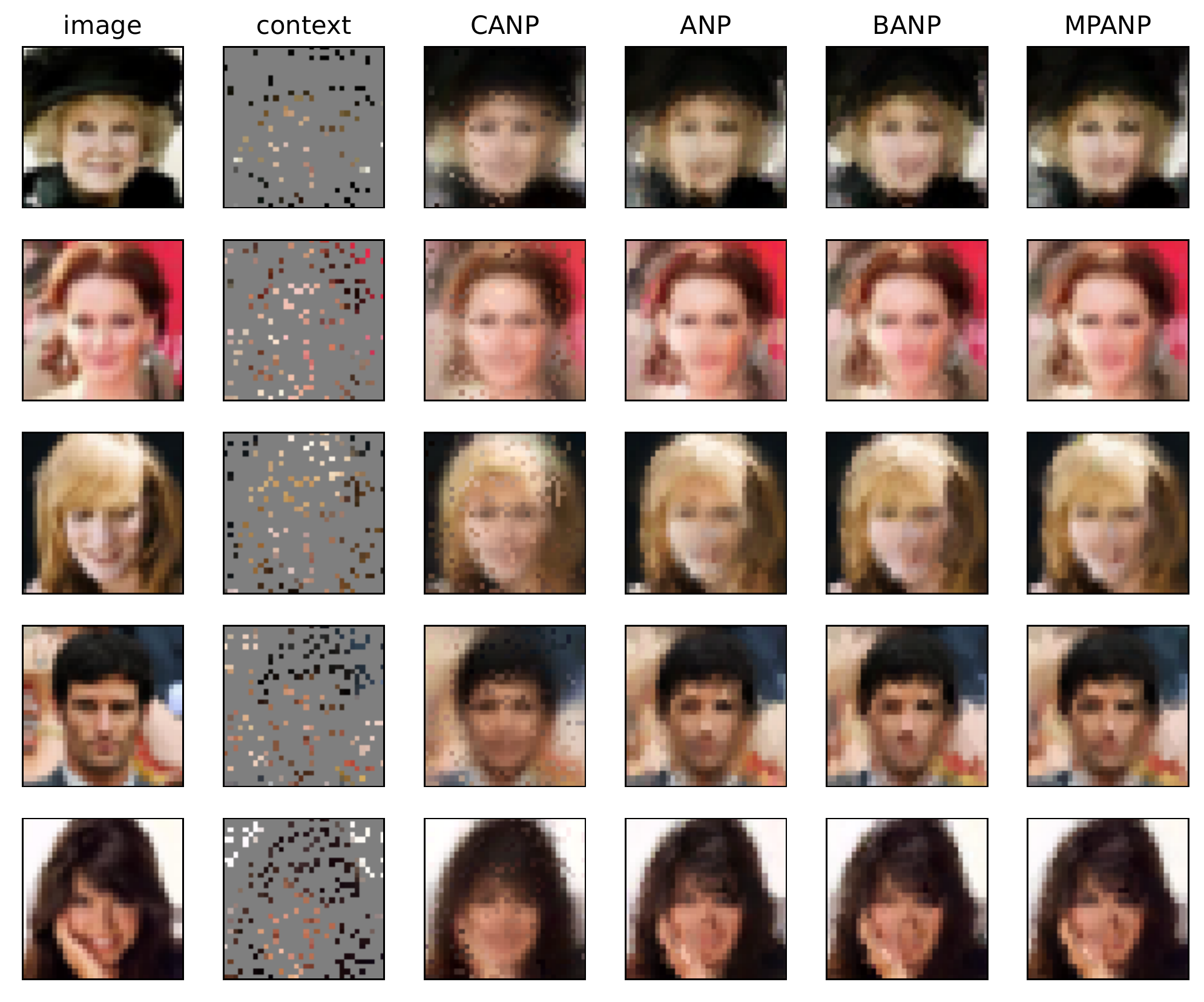}
    \caption{Predicted mean of image pixels by trained \gls{canp}, \gls{anp}, \gls{banp} and \gls{mpanp} model. (Column 1) Here we can see the 5 ground truth real image from the test dataset. (Column 2) The context set which given to the models. (Column 3-6) The predicted mean of image pixels by each models.}
    \label{figure/app_visualize_celeba}
\end{figure}

\paragraph{MNIST}
We provide some completed MNIST images in~\cref{figure/app_visualize_mnist}. It shows that both \gls{mpnp} and \gls{mpanp} successfully fill up the remaining parts of the image for a given context and capture the uncertainties as predictive variances.

\paragraph{CelebA}
We also present five examples from the CelebA dataset in~\cref{figure/app_visualize_celeba}. It shows that \gls{mpanp} provides perceptually reasonable predictions even for complex three-channel images.


\subsection{Bayesian Optimization}
\label{app:sec:additional_experiments:bo}

\begin{figure}
\centering
\includegraphics[width=0.49\linewidth]{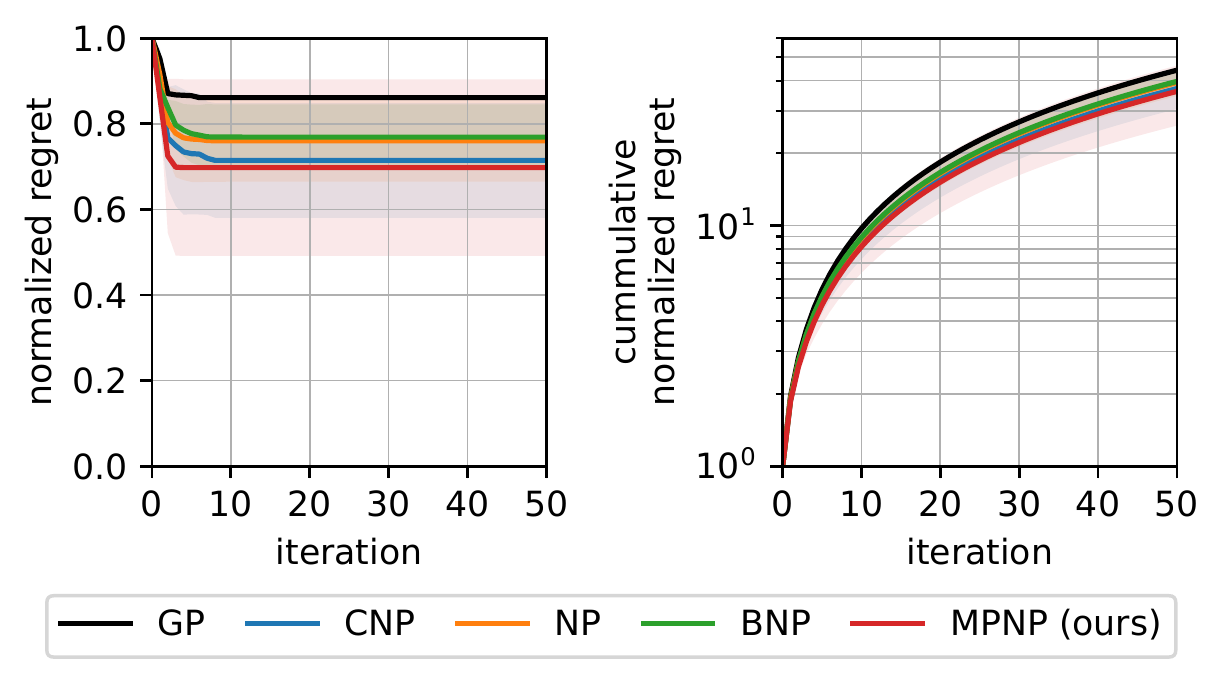}
\includegraphics[width=0.49\linewidth]{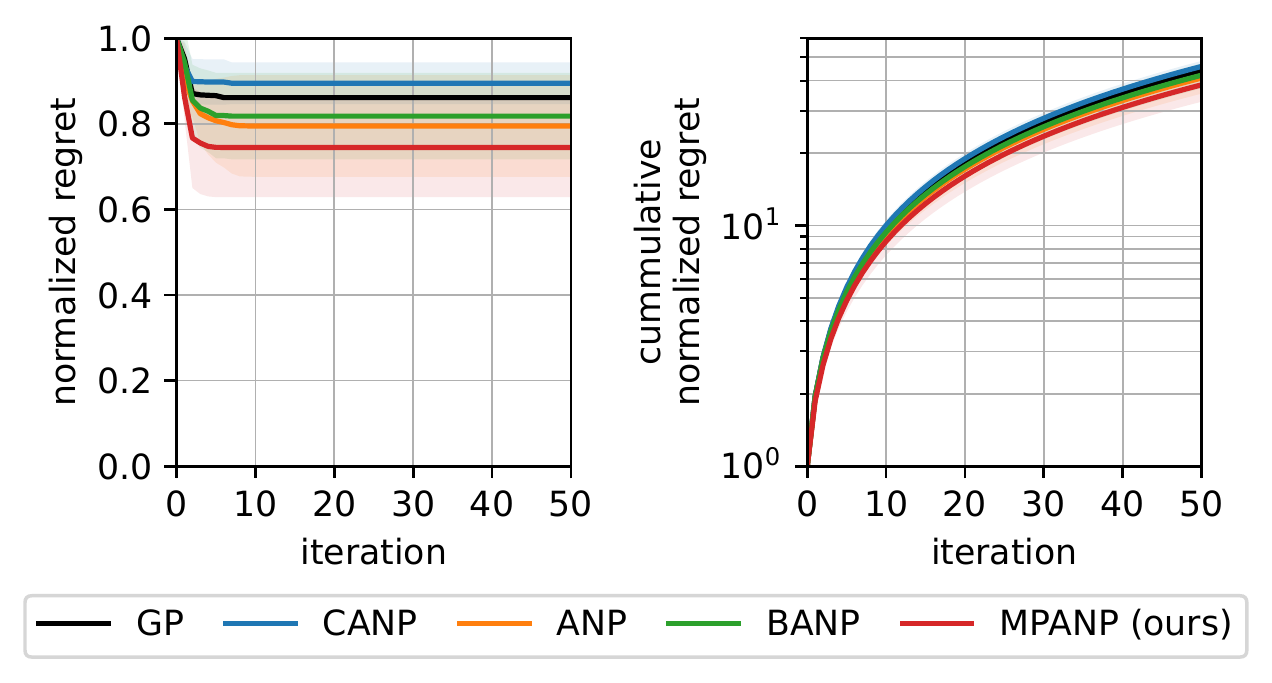}
\caption{Results for Bayesian optimization on~\citet{sobester2008engineering} function.}
\label{figure/app_bo_sob}
\end{figure}
\begin{figure}
\centering
\begin{subfigure}[b]{\textwidth}
\includegraphics[width=\linewidth]{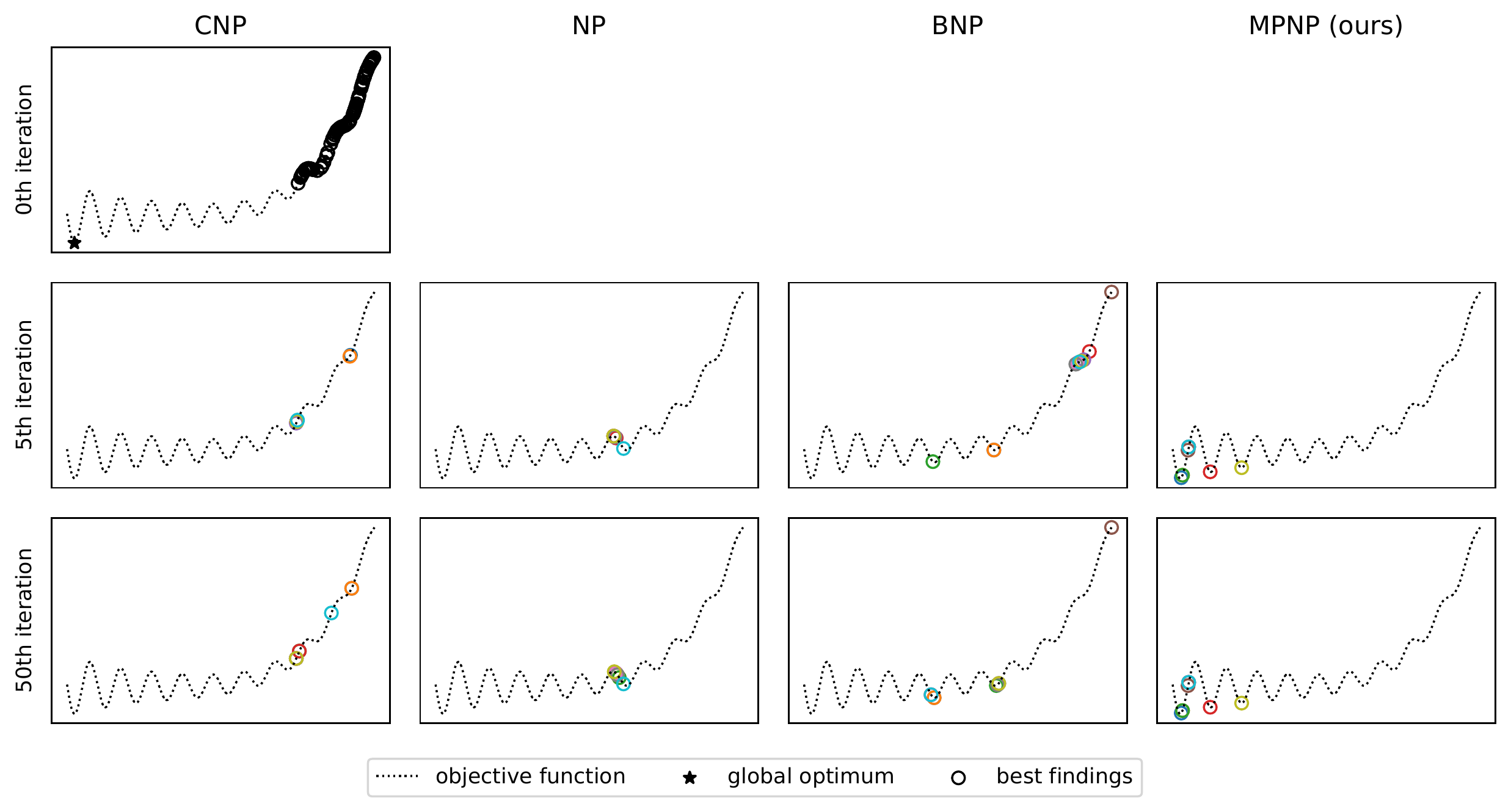}
\caption{\citet{gramacy2012cases} function}
\end{subfigure}
\bigskip

\begin{subfigure}[b]{\textwidth}
\includegraphics[width=\linewidth]{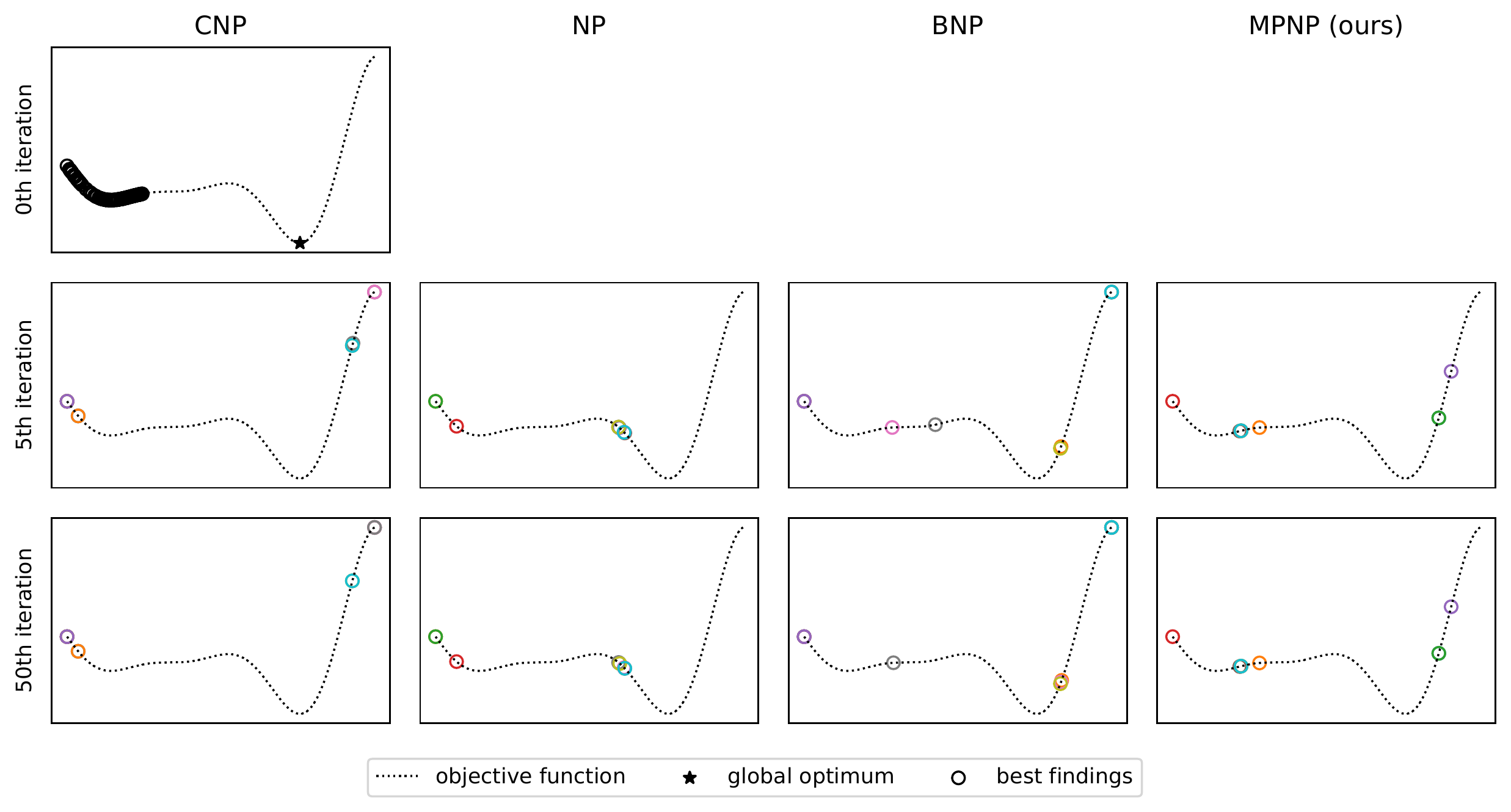}
\caption{\citet{sobester2008engineering} function}
\end{subfigure}
\caption{It depicts 10 solutions predicted by \gls{cnp}, \gls{np}, \gls{bnp}, and \gls{mpnp}. (a,b) Predicted results for \citet{gramacy2012cases} function and \citet{sobester2008engineering} function, respectively. (Row 1) Black circles indicate the whole initial points. (Row 2) It shows the 10 best solutions predicted by each models after the 5 iterations. (Row 3) It shows the 10 best solutions predicted by each models after the whole iterations.}
\label{figure/app_visualize_bo_np}
\end{figure}
\begin{figure}
\centering
\begin{subfigure}[b]{\textwidth}
\includegraphics[width=\linewidth]{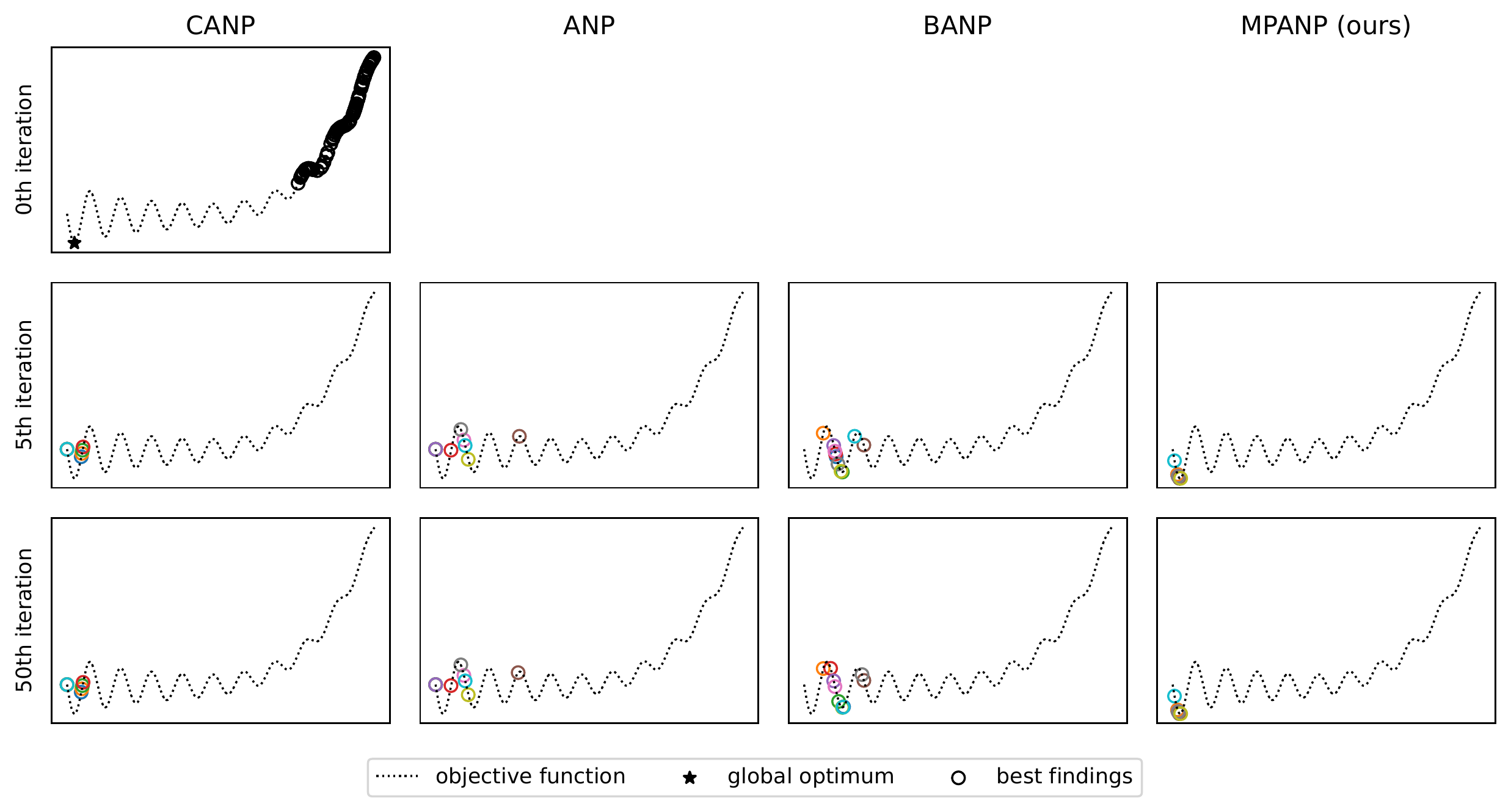}
\caption{\citet{gramacy2012cases} function}
\end{subfigure}
\bigskip

\begin{subfigure}[b]{\textwidth}
\includegraphics[width=\linewidth]{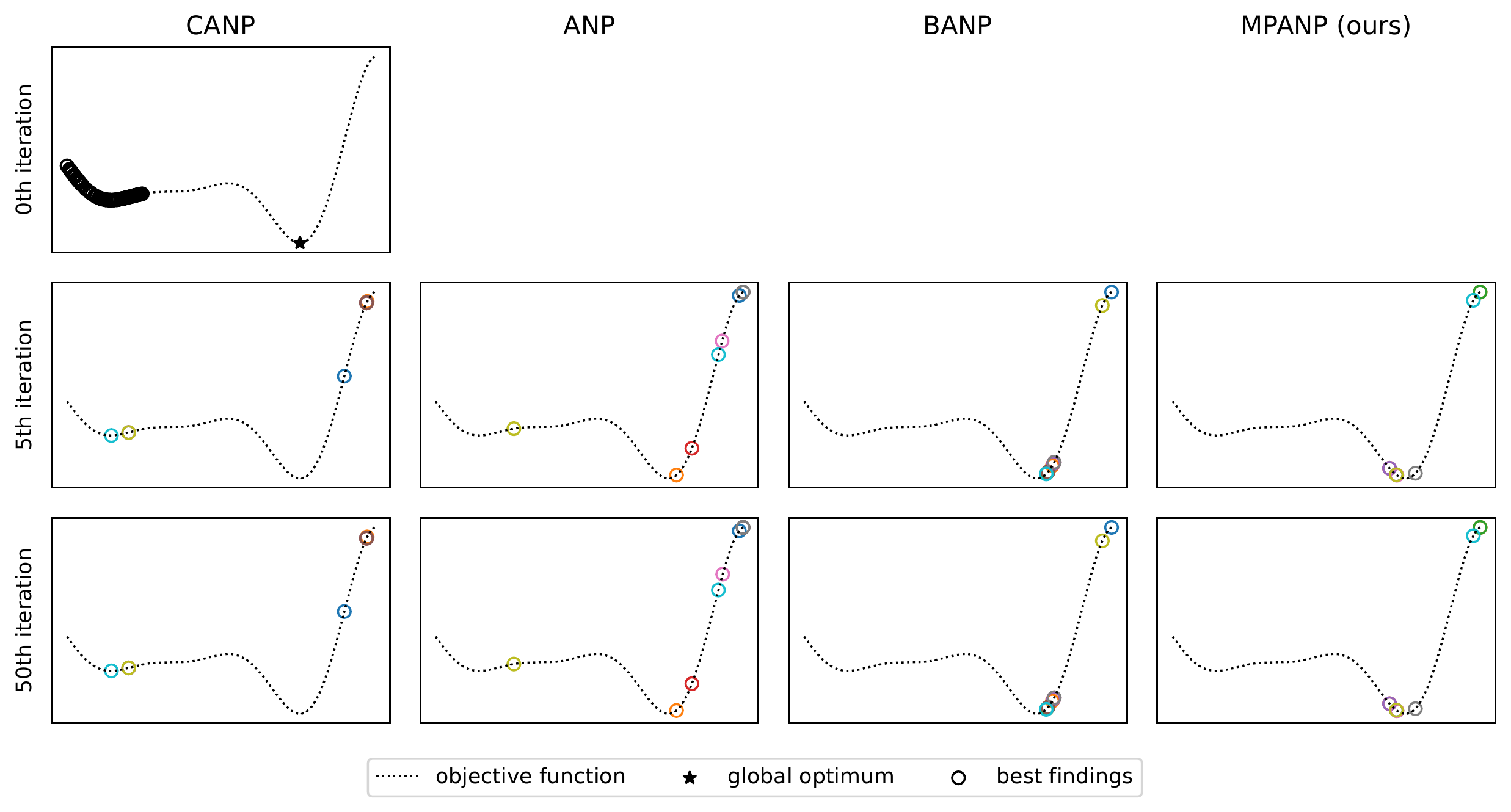}
\caption{\citet{sobester2008engineering} function}
\end{subfigure}
\caption{It depicts 10 solutions predicted by \gls{canp}, \gls{anp}, \gls{banp}, and \gls{mpanp}. (a,b) Predicted results for \citet{gramacy2012cases} function and \citet{sobester2008engineering} function, respectively. (Row 1) Black circles indicate the whole initial points. (Row 2) It shows the 10 best solutions predicted by each models after the 5 iterations. (Row 3) It shows the 10 best solutions predicted by each models after the whole iterations.}
\label{figure/app_visualize_bo_anp}
\end{figure}

We provide the results for Bayesian optimization on the~\citet{sobester2008engineering} function in~\cref{figure/app_bo_sob}. Our \glspl{mpnp} consistently outperform baselines as discussed in~\cref{main:sec:experiments:bo}. We also present the visual results for Bayesian optimization in~\cref{figure/app_visualize_bo_np,figure/app_visualize_bo_anp}.

\section{Experimental Details}
\label{app:sec:details}
We attached our code in supplementary material. Our codes used python libraries JAX~\citep{jax2018github}, Flax~\citep{flax2020github} and Optax~\citep{optax2020github}. 
These python libraries are available under the Apache-2.0 license\footnote{\href{https://www.apache.org/licenses/LICENSE-2.0}{https://www.apache.org/licenses/LICENSE-2.0}}.

We conducted all experiments on a single NVIDIA GeForce RTX 3090 GPU, except for the image completion tasks presented in~\cref{main:sec:experiments:imagecompletion}; we used 8 TPUv3 cores supported by TPU Research Cloud\footnote{\href{https://sites.research.google/trc/about/}{https://sites.research.google/trc/about/}} for the 2D image completion task. For optimization, we used Adam~\citep{kingma2015adam} optimizer with a cosine learning rate schedule. Unless specified, we selected the base learning rate from a grid of $\{5 \times 10^{-4.50}, 5 \times 10^{-4.25}, 5 \times 10^{-4.00}, 5 \times 10^{-3.75}, 5 \times 10^{-3.50}\}$ based on validation task log-likelihood.



\subsection{Evaluation metric}
Following \citet{le2018empirical},  for \gls{cnp} and \gls{canp}, which are deterministic models, we used the normalized predictive log-likelihood $\frac{1}{n}\sum_{i=1}^n\log p(y_i|x_i,Z_c)$. For other models, we used a approximation of the normalized predictive log-likelihood as:
\[
\frac{1}{n}\sum_{i=1}^n\log p(y_i|x_i,Z_c)\approx \frac{1}{n}\sum_{i=1}^n\log\frac{1}{K}\sum_{k=1}^K p(y_i|x_i,\theta^{(k)}),
\]
where $\theta^{k}$s are independent samples for $k\in[K]$.
\subsection{1D Regression}
To generate tasks $(Z, c)$, we first sample $x \iidsim \text{Unif}(-2, 2)$ and generate $Y$ using each kernel. We use RBF kernel $k(x, x') = s^2 \cdot \exp\left(\frac{-|| x - x' ||^2}{2\ell^2}\right)$, Matern $5/2$ kernel $k(x, x') = s^2 \cdot \left( 1 + \frac{\sqrt{5}d}{\ell} + \frac{5d^2}{3\ell^2} \right)$, and periodic kernel $k(x, x') = s^2 \cdot \exp\left( \frac{-2 \sin^2 (\pi||x-x'||^2 / p)}{\ell^2} \right)$ where all kernels use $s \sim \text{Unif}(0.1. 1.0)$, $\ell \sim \text{Unif}(0.1. 0.6)$, and $p \sim \text{Unif}(0.1. 0.5)$. To generate t-noise dataset, we use Student-$t$ with degree of freedom $2.1$ to sample noise $\epsilon \sim \gamma \cdot \calT(2.1)$ where $\gamma \sim \text{Unif}(0, 0.15)$. Then we add the noise to the curves generated from RBF kernel. We draw index set $|c| \sim \text{Unif}(3, 50 - 3)$ and $n - |c| \sim \text{Unif}(3, 50 - |c|)$ to maintain $\max |Z| \leq 50$. We use a batch size of $256$ for training.

\subsection{Image Completion}

We use the following datasets for image completion experiments.

\paragraph{MNIST}
We split MNIST~\citep{lecun1998gradient} train dataset into train set with 50,000 samples and validation set with 10,000 samples. We use whole 10,000 samples in test dataset as test set. We make $28 \times 28$ grids which both axes starting from $-0.5$ to $0.5$ to indicate the coordinate of pixels, and normalize pixel values into $[-0.5, 0.5]$. We use a batch size of $128$ for training.

\paragraph{SVHN}
We split SVHN~\citep{netzer2011reading} train dataset into train set with 58,600 samples and validation set with 14,657 samples. We use whole 26,032 samples in test dataset as test set. We make $32 \times 32$ grids which both axes starting from $-0.5$ to $0.5$ to indicate the coordinate of pixels, and normalize pixel values into $[-0.5, 0.5]$. We use a batch size of $128$ for training.

\paragraph{CelebA}
We use splits of CelebA~\citep{liu2015faceattributes} dataset as provided (162,770 train samples, 19,867 validation samples, 19,962 test samples). We crop $32 \times 32$ pixels of center of images. We make $32 \times 32$ grids which both axes starting from $-0.5$ to $0.5$ to indicate the coordinate of pixels, and normalize pixel values into $[-0.5, 0.5]$. We use a batch size of $128$ for training.


\subsection{Bayesian Optimization}
\label{app:sec:details:bo}

We use the following benchmark functions for Bayesian optimization experiments. Throughout the experiments, we adjust the function to  have the domain of $[-2.0, 2.0]$.
\paragraph{\citet{gramacy2012cases} function}
\[
f(x)=\frac{\sin(10\pi x)}{2x} + (x-1)^4,
\]
where $x\in [0.5,2.5]$ and a global optimum is at $x^\ast\approx 0.5486$.
\paragraph{\citet{sobester2008engineering} function}
\[
f(x)=(6x-2)^2\sin (12x-4),
\]
where $x\in [0,1]$ and a global optimum is at $x^\ast\approx0.7572$.

\section{Directly Generating Input Model}
\label{app:sec:directly_generating_input}
In this section, we present our model generating pseudo contexts directly in the input space.
We will present two kinds of model structure, i) directly generating pseudo context pair $(x,y)$ simultaneously by ISAB, ii) generating pseudo context data $x$ and $y$, sequentially. 
\subsection{Construction}
\paragraph{Generating pseudo context pair simultaneously.}
The generator of our first model which simultaneously generating pseudo context pair $(x',y')$, takes real context dataset $Z_c$ as input and outputs pseudo context dataset $Z'$. 
Here the generator is the one layer ISAB module.
Then we concatenate $Z_c$ and $Z'$ in order to treat this concatenated set as context dataset.
Then the encoder takes this concatenated context set as input.
And the others are the same with \gls{cnp} or \gls{canp}.
\paragraph{Sequentially generating pseudo context data \texorpdfstring{$x$}{x} and \texorpdfstring{$y$}{y}}
In this model, the generator takes real context dataset $Z_c$ as input and outputs only $x'$s of $Z'$. 
Here the generator is the one layer ISAB module with additional one linear layer.
Then we consider these $x'$s as our target dataset and find the mean and variance of $y'$ for each $x'$ by forwarding the model with context dataset $Z_c$ and target $x'$.
We sample $y'$ from the Gaussian distribution with mean and variance from the prior step.
We again concatenate $Z_c$ with $Z'$ and use them as context dataset.

\paragraph{Training}
Having directly generated a pseudo context set, we construct our empirical density as
\[
g_N(z) = \frac{1}{N}\bigg(\sum_{i\in c}\delta_{z_i}(z) + \sum_{i=1}^{N-|c|} \delta_{z'_i}(z)\bigg).
\]
Given $g_N$, we find the function parameter $\theta$ as
\[
\theta(g_N) := \argmin_\theta \int \ell(z, \theta) g_N(dz),
\]
where we simply choose $l(z,\theta):=-\log \calN (y|\mu_\theta(x),\sigma_\theta^2(x)I_{d_{out}})$.
In order to train the directly generating input model, which well approximate $\theta(g_N)$, we should construct different objective function from \cref{eq:mpnp_term_whole} because we can compute the exact $\int \ell(z, \theta) g_N(dz)$, unlike the feature generating model.
First, we approximate the marginal likelihood which is,
\[
\label{eq:direct1}
\log p(Y|X, Z_c) \approx \log \Bigg[ \frac{1}{K} \sum_{k=1}^K \exp\bigg(-\sum_{i\in [n]}\ell(z_i, \tilde\theta(Z_c\cup Z'^{(k)}))\bigg)
\Bigg] := -\calL_\text{marg}(\tau,\phi),
\]
where $Z'^{(1)},\dots, Z'^{(K)}\iidsim p(Z'|Z_c;\phi_\text{pred})$.
\cref{eq:direct1} is the same training object with \cref{eq:mpnp_term1}.
As we mentioned in \cref{main:subsec:training}, if we are given sufficiently well approximated $\Tilde{\theta}(Z_c\cup Z^{'(K)})$ then this objective would be suffice. However only with \cref{eq:direct1}, we cannot train the encoder to properly amortize the parameter construction process \cref{eq:recovering_theta}. To overcome this issue, we use $\int \ell(z, \theta) g_N(dz)$ as our second training objective which is,
\[
\label{eq:direct2}
\frac{1}{K}\sum_{k=1}^K\int \ell(z,\theta)g_N^{(k)}(dz) = \frac{1}{K}\sum_{k=1}^K\sum_{z\in Z_c\cup Z^{'(k)}} \Big(-\ell \big(z, \tilde{\theta}(Z_c\cup Z^{'(k)}\big)\Big)
:= \calL_\text{amort}(\tau,\phi).
\]
Combining these two functions, our loss function for the direct \gls{mpnp} is then
\[
\bbE_{\tau}[\calL(\tau,\phi)] = \bbE_\tau[\calL_\text{marg}(\tau,\phi) + \calL_\text{amort}(\tau,\phi)].
\]

\subsection{Sample}
\begin{table}[t]
\centering

\caption{Test results for 1D regression tasks on RBF. `Context' and `Target' respectively denote context and target log-likelihood values, and `Task' denotes the task log-likelihood. All values are averaged over four seeds.}
\label{table/app_gp_inf_direct}
\begin{tabular}{lrrr}
\toprule
& \multicolumn{3}{c}{RBF}  \\
\cmidrule(lr){2-4}
Model & Context & Target & Task  \\
\midrule
CNP & 1.096$\spm{0.023}$ &  0.515$\spm{0.018}$ &  0.796$\spm{0.020}$ \\
NP  & 1.022$\spm{0.005}$ &  0.498$\spm{0.003}$ &  0.748$\spm{0.004}$ \\
BNP & 1.112$\spm{0.003}$ &  0.588$\spm{0.004}$ &  0.841$\spm{0.003}$ \\
\textBF{MPNP (ours)} & 
\textBF{1.189}$\spm{0.005}$ & \textBF{0.675}$\spm{0.003}$ & \textBF{0.911}$\spm{0.003}$ \\
\textBF{MPNP DSI(ours)} & 
1.120$\spm{0.007}$ & 0.551$\spm{0.006}$ & 0.822$\spm{0.007}$ \\
\textBF{MPNP DSE(ours)} & 
1.121$\spm{0.007}$ & 0.555$\spm{0.006}$ & 0.824$\spm{0.007}$ \\
\midrule
CANP & 
 1.304$\spm{0.027}$ &  0.847$\spm{0.005}$ &  1.036$\spm{0.020}$  \\
ANP & 
 \textBF{1.380}$\spm{0.000}$ &  0.850$\spm{0.007}$ &  1.090$\spm{0.003}$  \\
BANP & 
 \textBF{1.380}$\spm{0.000}$ &  0.846$\spm{0.001}$ &  1.088$\spm{0.000}$  \\
\textBF{MPANP (ours)} & 
 1.379$\spm{0.000}$ &  \textBF{0.881}$\spm{0.003}$ &  \textBF{1.102}$\spm{0.001}$ \\
 \textBF{MPANP DSI(ours)} & 
 \textBF{1.380}$\spm{0.000}$ &  0.796$\spm{0.013}$ &  1.069$\spm{0.005}$ \\
  \textBF{MPANP DSE(ours)} & 
 1.380$\spm{0.000}$ &  0.783$\spm{0.014}$ &  1.064$\spm{0.005}$ \\
\bottomrule
\end{tabular}
\end{table}

\begin{figure}[t]
    \centering
    \includegraphics[width = 0.49\textwidth]{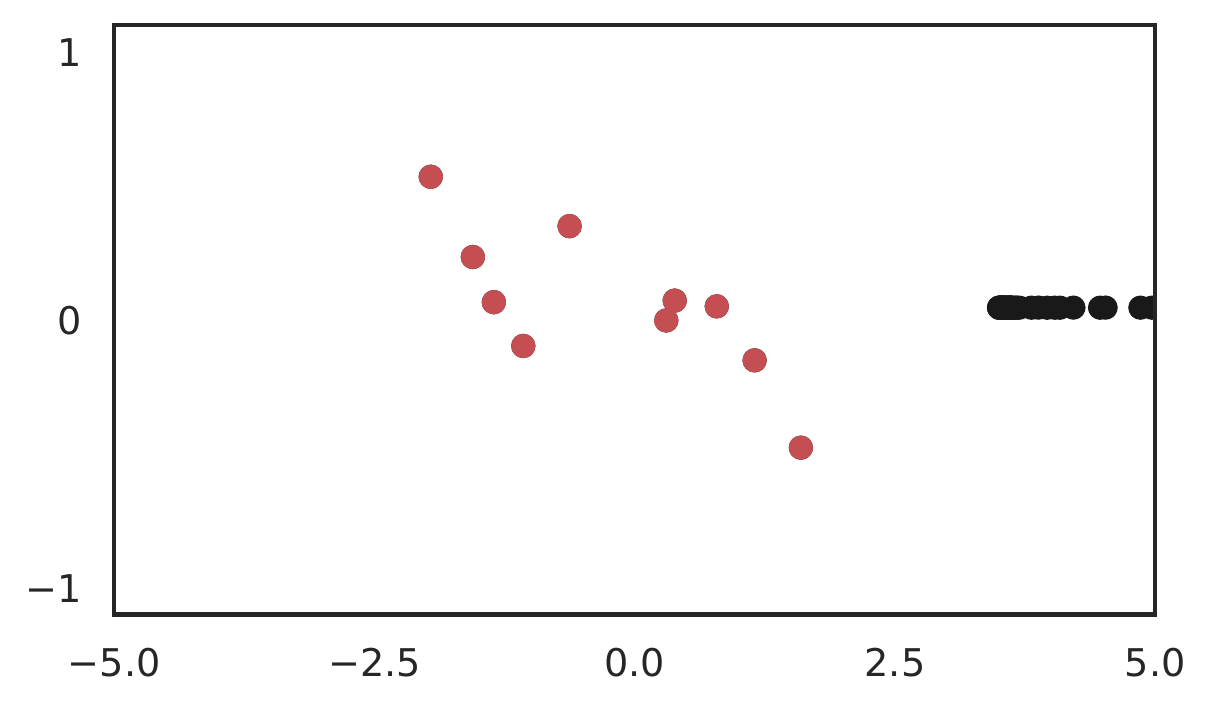}
    \includegraphics[width = 0.49\textwidth]{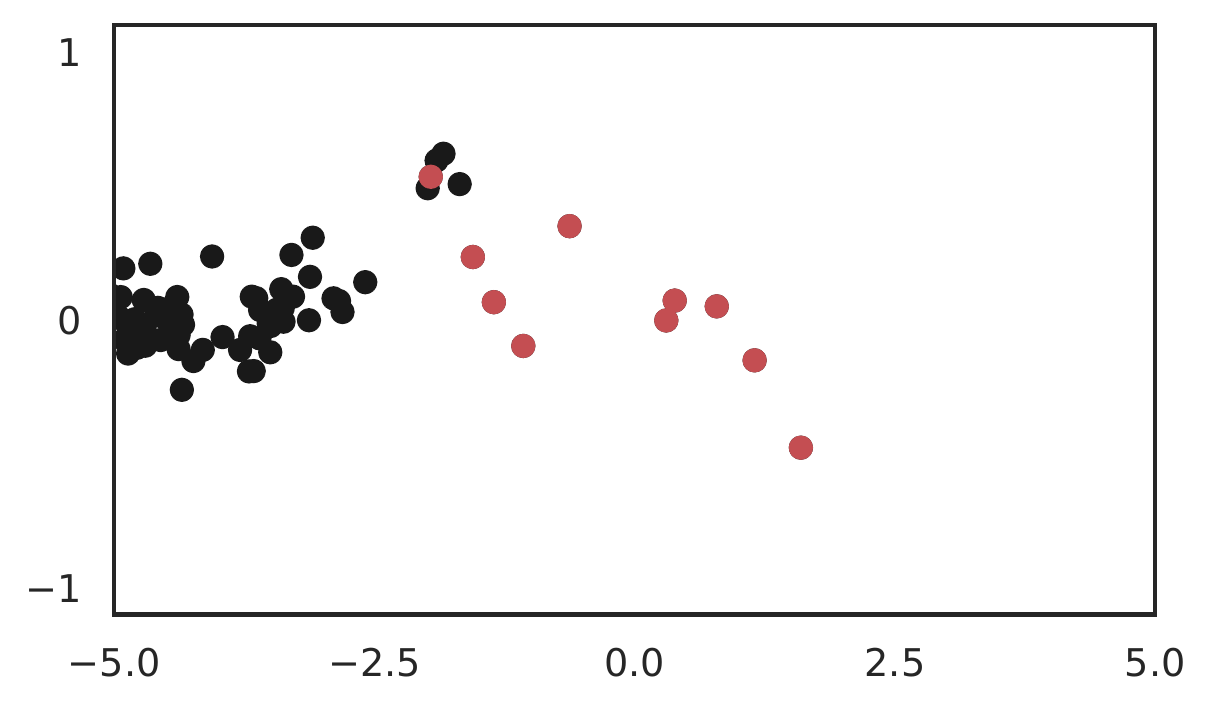}
    \caption{It shows generated pseudo context dataset of direct \gls{mpanp} for 1D regression task with RBF kernel. The red dots are true context points sampled from \gls{gp} with RBF kernel, and the black dots are generated pseudo context points. (Left) Results from simultaneously generating pseudo context pair \gls{mpanp} model. (Right) Results from sequentially generating pseudo context data \gls{mpanp} model.}
    \label{fig:direct_sample}
\end{figure}
\begin{figure}[t]
    \centering
    \includegraphics[width = 0.49\textwidth]{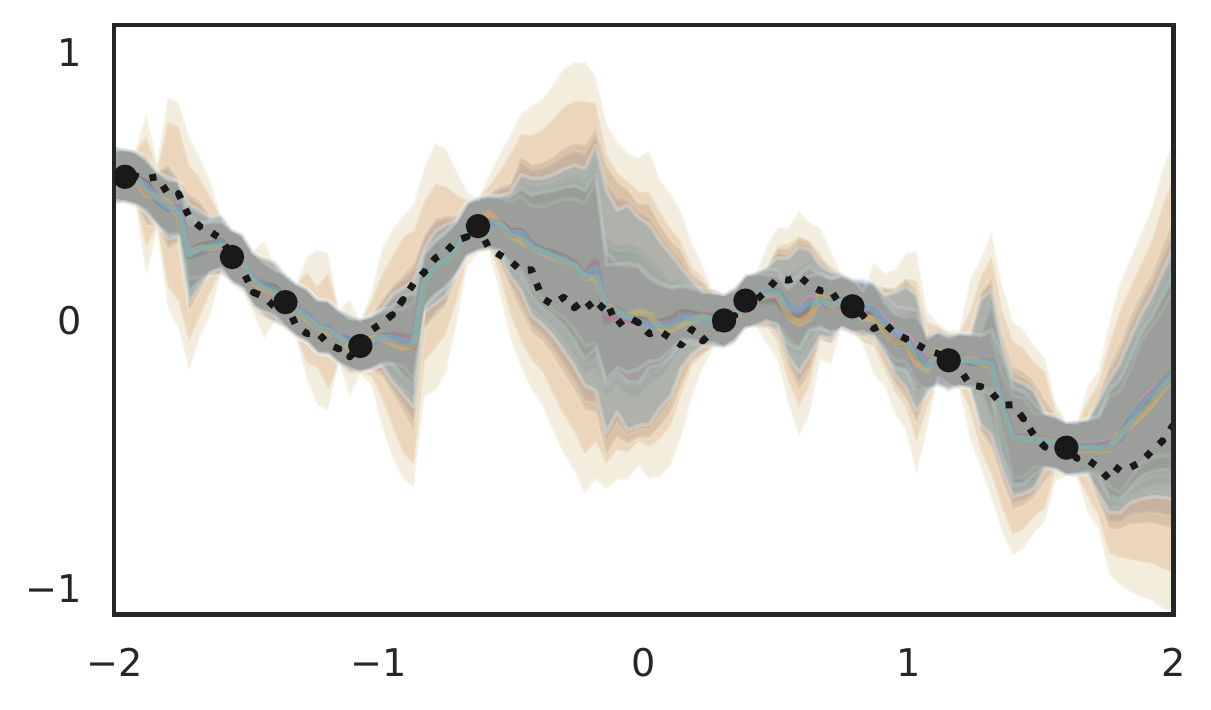}
    \includegraphics[width = 0.49\textwidth]{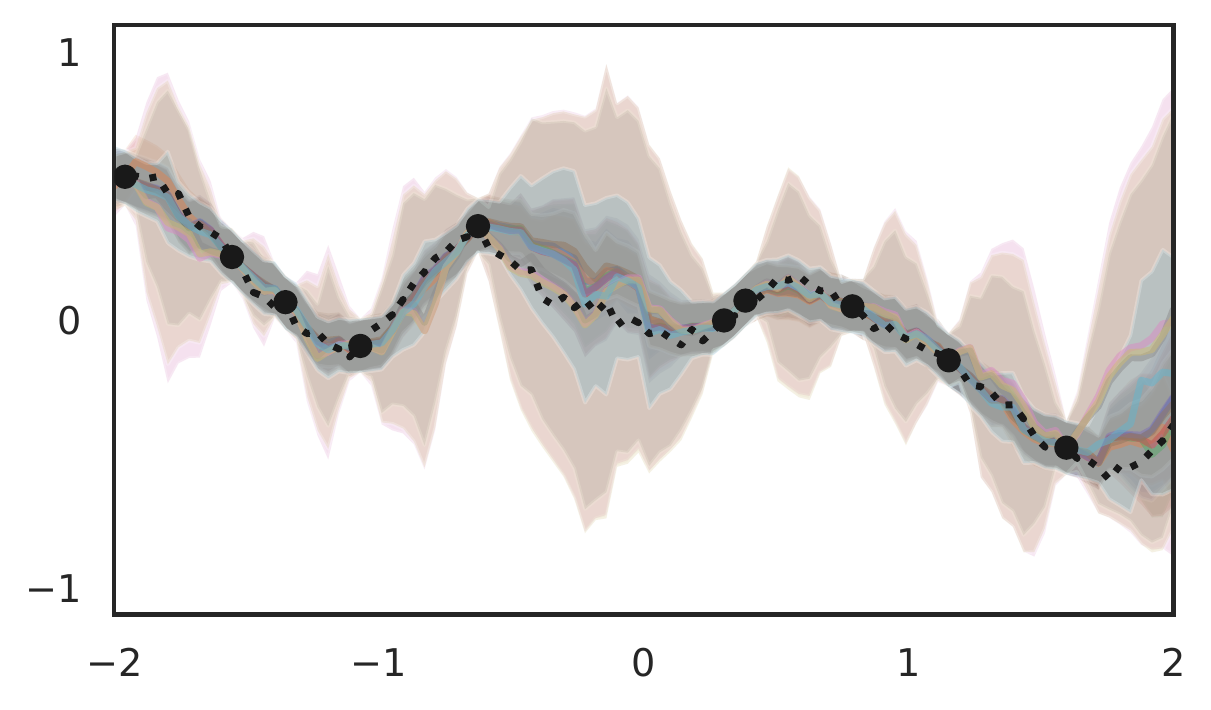}
    \caption{It shows posterior samples of direct \gls{mpanp} for 1D regression task with RBF kernel. The black dashed line is a function sampled from \gls{gp} with RBF kernel, and the black dots are context points. We visualized decoded mean and standard deviation with colored lines and areas. (Left) Results from simultaneously generating pseudo context pair \gls{mpanp} model. (Right) Results from sequentially generating pseudo context data \gls{mpanp} model.}
    \label{fig:direct_decode}
\end{figure}

In this section, we presents how the directly generating input model actually samples the pseudo context datasets. 

In \cref{fig:direct_sample}, we report generated pseudo context datasets and posterior samples from two different cases of directly generating input models for 1D regression task with RBF kernel. Here we can see that the generator samples pseudo context datasets far from the real context dataset. This phenomenon occurs because the generator learns to generate meaningless inputs ignored by the decoder.
In \cref{fig:direct_decode}, we report how two different directly generating \glspl{mpanp} predict posterior samples for 1D regression task with RBF kernel. 
Although directly generated pseudo context dataset are a bit far from context dataset, our model still well capture the functional uncertainty in this case.
We report the test results for 1D regression tasks on RBF for two directly generating models in \cref{table/app_gp_inf_direct}.
DSI and DSE indicate simultaneously generating models and sequentially generating models, respectively.
\cref{table/app_gp_inf_direct} shows that our directly generating models still outperform \gls{cnp} and \gls{canp} in the perspective of log-likelihood.


\end{document}